\documentclass[10pt,twocolumn,letterpaper]{article}

\usepackage{cvpr}              

\usepackage{graphicx}
\usepackage{amsmath}
\usepackage{amssymb}
\usepackage{booktabs}

\usepackage[pagebackref,breaklinks,colorlinks]{hyperref}

\newcommand{\norm}[1]{\left\lVert#1\right\rVert}

\usepackage{titling}
\date{}
\predate{}
\postdate{}
\usepackage[resetlabels]{multibib}
\newcites{Main}{References}%
\newcites{Supp}{References}%

\usepackage[capitalize]{cleveref}
\crefname{section}{Sec.}{Secs.}
\Crefname{section}{Section}{Sections}
\Crefname{table}{Table}{Tables}
\crefname{table}{Tab.}{Tabs.}

\usepackage{wrapfig}
\renewcommand\footnotemark{}

\def\cvprPaperID{2430}
\def\confName{CVPR}
\def\confYear{2023}

\begin{document}

\title{\vspace{-0.2em} \Large \textbf{DiffPose: Toward More Reliable 3D Pose Estimation}   \vspace{0.3em}}

\author{Jia Gong\textsuperscript{1\dag}\thanks{ \dag~Equal contribution;~~\S~Currently at Meta;~~\ddag~Corresponding author}
~~~ Lin Geng Foo\textsuperscript{1\dag}
~~~ Zhipeng Fan\textsuperscript{2\S}
~~~ Qiuhong Ke\textsuperscript{3}
~~~ Hossein Rahmani\textsuperscript{4} ~~~ Jun Liu\textsuperscript{1\ddag} \\
\textsuperscript{1}Singapore University of Technology and Design ~~ \\
\textsuperscript{2}New York University ~~ 
\textsuperscript{3}Monash University ~~ 
\textsuperscript{4}Lancaster University\\
{\tt\small \{jia\_gong,lingeng\_foo\}@mymail.sutd.edu.sg, zf606@nyu.edu, qiuhong.ke@monash.edu, }\\ 
{\tt\small h.rahmani@lancaster.ac.uk, jun\_liu@sutd.edu.sg } \\
}

\maketitle

\begin{abstract}
Monocular 3D human pose estimation is quite challenging due to the inherent ambiguity and occlusion, which often lead to high uncertainty and indeterminacy. On the other hand, diffusion models have recently emerged as an effective tool for generating high-quality images from noise. Inspired by their capability, we explore a novel pose estimation framework (DiffPose) that formulates 3D pose estimation as a reverse diffusion process. We incorporate novel designs into our DiffPose to facilitate the diffusion process for 3D pose estimation: a pose-specific initialization of pose uncertainty distributions, a Gaussian Mixture Model-based forward diffusion process, and a context-conditioned reverse diffusion process. Our proposed DiffPose significantly outperforms existing methods on the widely used pose estimation benchmarks Human3.6M and MPI-INF-3DHP. Project page: https://gongjia0208.github.io/Diffpose/.
\end{abstract}

\section{Introduction}
3D human pose estimation, which aims to predict the 3D coordinates of human joints from images or videos, is an important task with a wide range of applications, including augmented reality~\citeMain{chessa2019grasping}, sign language translation~\citeMain{liang2020multi} and human-robot interaction~\citeMain{sridhar2015investigating},
attracting a lot of attention in recent years \citeMain{zhaoCVPR19semantic,liu2020comprehensive,xu2021graph,zhao2022graformer}. 
Generally, the mainstream approach is to conduct 3D pose estimation in two stages: the 2D pose is first obtained with a 2D pose detector, and then 2D-to-3D lifting is performed (where the lifting process is the primary aspect that most recent works \citeMain{pavllo20193d,cai2019exploiting,zheng20213d,li2022mhformer,khirodkar2021multi,li2019generating,foo2023unified} focus on).
Yet, despite the considerable progress, monocular 3D pose estimation still remains challenging.
In particular, it can  be difficult to accurately predict 3D pose from monocular data due to many challenges, including the inherent depth ambiguity and the potential occlusion, which often lead to high indeterminacy and uncertainty.

\begin{figure}
  \centering
  \includegraphics[width=1\linewidth]{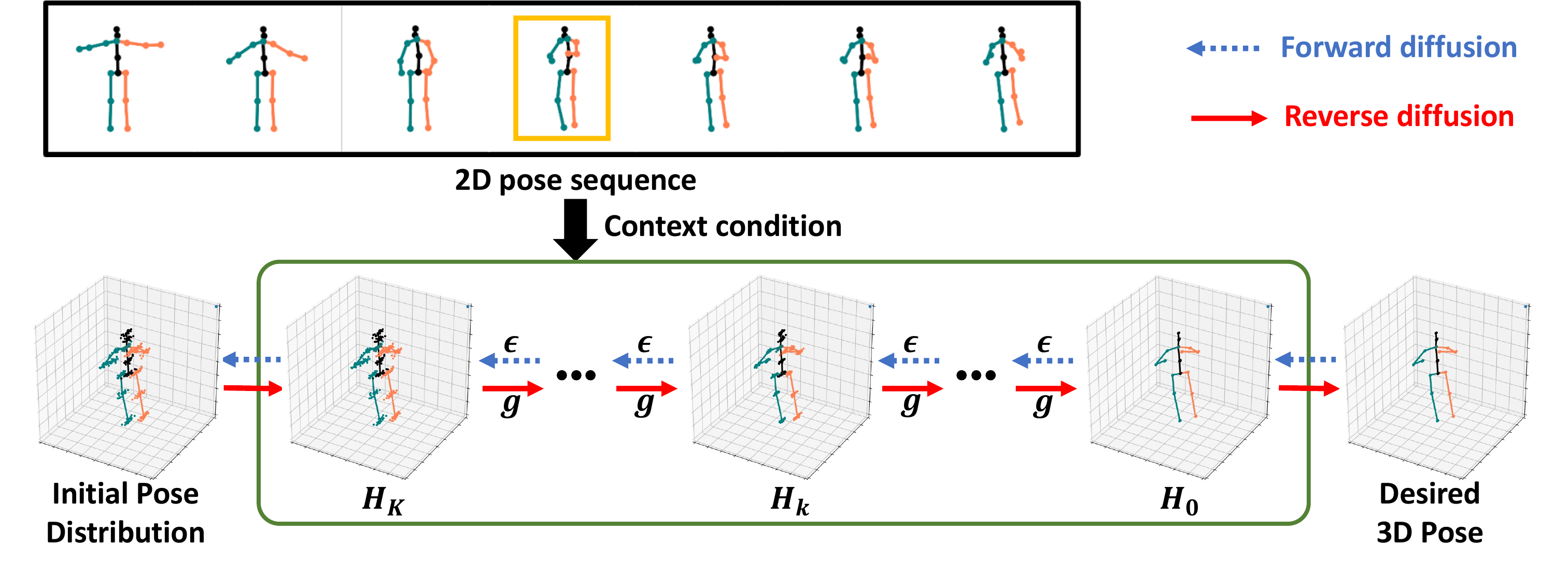}
  \vspace{-8mm}
  \captionof{figure}{
Overview of our DiffPose framework.
In the forward process (denoted with {blue dotted arrows}), we gradually diffuse a ``ground truth" 3D pose distribution $H_0$ with low indeterminacy towards a 3D pose distribution with high uncertainty $H_K$ by adding noise $\epsilon$ at every step, which generates intermediate distributions to guide model training. 
Before the reverse process, we first initialize the indeterminate 3D pose distribution $H_K$ from the input.
Then, during the reverse process (denoted with \textcolor{red}{red solid arrows}), we use the diffusion model $g$, conditioned on the context information from 2D pose sequence, to progressively transform $H_K$ into a 3D pose distribution $H_0$ with low indeterminacy.
}
\label{fig: Pose Reverse Diffusion}
\vspace{-5 mm}
\end{figure}

On the other hand, diffusion models \citeMain{ho2020denoising,song2021denoising} have recently become popular as an effective way to generate high-quality images \citeMain{rombach2022high}.
Generally, diffusion models are capable of generating samples that match a specified data distribution (e.g., natural images) from random (indeterminate) noise through multiple steps where the noise is progressively removed  \citeMain{ho2020denoising,song2021denoising}.
Intuitively, such a paradigm of progressive denoising helps to break down the large gap between distributions (from a highly uncertain one to a determinate one) into smaller intermediate steps \citeMain{song2019generative} and thus successfully helps the model to converge towards smoothly generating samples from the target data distribution.

Inspired by the strong capability of diffusion models to generate realistic samples even from a starting point with high uncertainty (e.g., random noise), here we aim to tackle 3D pose estimation, 
which also involves handling uncertainty and indeterminacy (of 3D poses), with diffusion models.
In this paper, we propose \textbf{DiffPose}, a novel framework that represents a new brand of diffusion-based 3D pose estimation approach, which also follows the mainstream two-stage pipeline.
In short, DiffPose models the 3D pose estimation procedure as a reverse diffusion process, where we progressively transform a 3D pose distribution with high uncertainty and indeterminacy towards a 3D pose with low uncertainty.

Intuitively, we can consider the determinate ground truth 3D pose as particles in the context of thermodynamics, where particles can be neatly gathered and form a clear pose with low indeterminacy at the start; then eventually these particles stochastically spread over the space, leading to high indeterminacy.
This process of particles evolving from low indeterminacy to high indeterminacy is the \emph{forward diffusion process}.
The pose estimation task aims to perform precisely the opposite of this process, i.e., \textit{the reverse diffusion process}. 
We receive an initial 2D pose that is indeterminate and uncertain in 3D space, and we want to shed the indeterminacy to obtain a determinate 3D pose distribution containing high-quality solutions.

Overall, our DiffPose framework consists of two opposite processes: the \textit{forward process} and the \textit{reverse process}, as shown in Fig. \ref{fig: Pose Reverse Diffusion}.
In short, the forward process generates supervisory signals of intermediate distributions for training purposes, while the reverse process is a key part of our 3D pose estimation pipeline that is used for both training and testing. 
Specifically, in the forward process, we gradually diffuse a ``ground truth" 3D pose distribution $H_0$ with low indeterminacy towards a 3D pose distribution with high indeterminacy that resembles the 3D pose's underlying uncertainty distribution $H_K$. We obtain samples from the intermediate distributions along the way, which are used during training as step-by-step supervisory signals for our diffusion model $g$.
To start the reverse process, we first initialize the indeterminate 3D pose distribution ($H_K$) according to the underlying uncertainty of the 3D pose. Then, our diffusion model $g$ is used in the reverse process to progressively transform $H_K$ into a 3D pose distribution with low indeterminacy ($H_0$).
The diffusion model $g$ is optimized using the samples from intermediate distributions (generated in the forward process), which guide it to smoothly transform the indeterminate distribution $H_K$ into accurate predictions.

However, there are several challenges in the above forward and reverse process. 
Firstly, in 3D pose estimation, we start the reverse diffusion process from an estimated 2D pose which has high uncertainty in 3D space, instead of starting from random noise like in existing image generation diffusion models \citeMain{ho2020denoising,song2021denoising}.
This is a significant difference, as it means that the underlying uncertainty distribution of each 3D pose can differ.
Thus, we cannot design the output of the forward diffusion steps to converge to the same Gaussian noise like in previous image generation diffusion works \citeMain{ho2020denoising,song2021denoising}. 
Moreover, the uncertainty distribution of 3D poses can be irregular and complicated, making it hard to characterize via a single Gaussian distribution.
Lastly, it can be difficult to perform accurate 3D pose estimation with just $H_K$ as input. 
This is because our aim is not just to generate any realistic 3D pose, but rather to predict accurate 3D poses corresponding to our estimated 2D poses, which often requires more context information to achieve.

To address these challenges, we introduce several novel designs in our DiffPose.
{Firstly, we initialize the indeterminate 3D pose distribution $H_K$ based on extracted heatmaps, which captures the underlying uncertainty of the desired 3D pose.}
Secondly, during forward diffusion, to generate the indeterminate 3D pose distributions that eventually (after $K$ steps) resemble $H_K$, we add noise to the ground truth 3D pose distribution $H_0$, where the noise is modeled by a Gaussian Mixture Model (GMM) that characterizes the uncertainty distribution $H_K$.
Thirdly, the reverse diffusion process is conditioned on context information from the input video or frame in order to better leverage the spatial-temporal relationship between frames and joints.
{Then, to effectively use the context information and perform the progressive denoising to obtain accurate 3D poses, we design a GCN-based diffusion model $g$.}

The contributions of this paper are threefold: 
(i) We propose DiffPose, a novel framework 
which represents a new brand of method with the diffusion architecture for 3D pose estimation, which can naturally handle the indeterminacy and uncertainty of 3D poses.
(ii) We propose various designs to facilitate 3D pose estimation, including the initialization of 3D pose distribution, a GMM-based forward diffusion process and a conditional reverse diffusion process.
(iii) DiffPose achieves state-of-the-art performance on two widely used human pose estimation benchmarks.

\section{Related Work}
 \textbf{3D Human Pose Estimation.}
Existing monocular 3D pose estimation methods can roughly be categorized into two groups: frame-based methods and video-based ones.
\textit{Frame-based methods} predict the 3D pose from a single RGB image. Some works \citeMain{pavlakos2017coarse, sun2018integral, fan2021motion, fan2020adaptive, park20163d, foo2023system} use Convolutional Neural Networks (CNNs) to output a human pose from the RGB image, while many works \citeMain{martinez2017simple,zhao2019semantic,zhao2022graformer,xu2021graph} first detect the 2D pose and then use it to regress the 3D pose.
On the other hand, \textit{video-based methods} tend to exploit temporal dependencies between frames in the video clip.
Most video-based methods \citeMain{pavllo20193d,chen2021anatomy,shan2021improving,cai2019exploiting,ci2019optimizing,zheng20213d,shan2022p,hu2021conditional,xu2020deep,wang2020motion,foo2023unified} extract 2D pose sequences from the input video clip via a 2D pose detector, and focus on distilling the crucial spatial-temporal information from these 2D pose sequences for 3D pose estimation.
To encode spatial-temporal information, existing works explore CNN-based frameworks with temporal convolutions \citeMain{pavllo20193d,chen2021anatomy}, GCNs \citeMain{cai2019exploiting,ci2019optimizing}, or Transformers \citeMain{zheng20213d,shan2022p}. 
Notably, several works \citeMain{li2022mhformer,sharma2019monocular,li2019generating} aim to alleviate the uncertainty and indeterminacy in 3D pose estimation by designing models that can generate multiple hypothesis solutions from a single input.
Different from all the aforementioned works, DiffPose is formulated as a \textit{distribution-to-distribution} transformation process, where we train a diffusion model to smoothly denoise from the indeterminate pose distribution to a pose distribution with low indeterminacy.
By framing the 3D pose estimation procedure as a reverse diffusion process, DiffPose can naturally handle the indeterminacy and uncertainty for 3D pose estimation.

\textbf{Denoising Diffusion Probabilistic Models (DDPMs).}
DDPMs (called diffusion models for short) have emerged as an effective approach to learn a data distribution that is straightforward to sample from. 
Introduced by Sohl-Dickstein et al. \citeMain{sohl2015deep} for image generation, DDPMs have been further simplified and accelerated \citeMain{ho2020denoising,song2021denoising}, 
{and enhanced \citeMain{nachmani2021non,Zhao_2023_arxiv_watermark_dm,austin2021structured,nichol2021improved} in recent years.}
Previous works have explored applying diffusion models to various generation tasks, including image inpainting \citeMain{lugmayr2022repaint} and text generation \citeMain{li2022diffusion}.
Here, we explore using diffusion models to tackle 3D pose estimation with our DiffPose framework.
Unlike these generation tasks \citeMain{lugmayr2022repaint,li2022diffusion} that often start the generation process from random noise, our pose estimation process starts from an estimated 2D pose with uncertainty and indeterminacy in 3D space, where the uncertainty distribution differs for each pose and can also be irregular and difficult to characterize.
We also design a GCN-based architecture as our diffusion model $g$, and condition it on spatial-temporal context information to aid the reverse diffusion process and obtain accurate 3D poses.

\section{Background on Diffusion Models}
\label{sec:background}
Diffusion models \citeMain{ho2020denoising,song2021denoising} are a class of probabilistic generative models that learn to transform noise $h_{K} \sim \mathcal{N}(\mathbf{0},\mathbf{I})$ to a sample $h_0$ by recurrently denoising $h_{K}$, i.e., $(h_K \rightarrow h_{K-1} \rightarrow ... \rightarrow h_{0})$. This denoising process is called $\emph{reverse diffusion}$. 
Conversely, the process $(h_0 \rightarrow h_{1} \rightarrow ... \rightarrow h_{K})$ is called $\emph{forward diffusion}$.

To allow the diffusion model to learn the reverse diffusion process, a set of intermediate noisy samples $\{h_k\}^{K-1}_{k=1}$ are needed to bridge the source sample $h_0$ and the Gaussian noise $h_{K}$.
Specifically, \emph{forward diffusion} is conducted to generate these samples, where the posterior distribution $q(h_{1:K}|h_0)$ from $h_{0}$ to $h_{K}$ is formulated as:

\begin{small}
\setlength{\abovedisplayskip}{4pt}
\label{eq: original diffusion}
\begin{align}
q(h_{1:K}|h_{0}) &:=  \prod_{k=1}^{K} q(h_{k}|h_{k-1})  \\
q(h_{k}|h_{k-1}) &:= \mathcal{N}_{pdf} \big(h_k \big| \sqrt{\frac{\alpha _k}{\alpha _{k-1}}}h_{k-1},(1-\frac{\alpha _k}{\alpha _{k-1}})\mathbf{I} \big),  
\end{align}
\end{small}
where $\mathcal{N}_{pdf}(h_k| \cdot)$ refers to the likelihood of sampling $h_k$ conditioned on the given parameters, and $\alpha _{1:K} \in (0,1]^K$ is a fixed decreasing sequence that controls the noise scaling at each diffusion step.
Using the known statistical results for the combination of Gaussian distributions,
the posterior for the diffusion process to step $k$ can be formulated as:
\begin{small}
\vspace{-1mm}
\setlength{\abovedisplayskip}{4pt}
\begin{align}
\label{eq: original diffusion q function} 
q(h_k|h_0) :=& \int q(h_{1:k}|h_0) \text{d} h_{1:k-1} \notag \\
=& \mathcal{N}_{pdf}(h_k | \sqrt{\alpha _k}h_0, (1-\alpha _k)\mathbf{I}).
\end{align}
\end{small}

Thus, $h_k$ can be expressed as a linear combination of the source sample $h_0$ and a noise variable $\epsilon$, where each element of $\epsilon$ is sampled from $\mathcal{N}(0,1)$, as follows:
\begin{small}
\setlength{\abovedisplayskip}{4pt}
\setlength{\belowdisplayskip}{4pt}
\begin{align}
\label{eq: original diffusion linear function} 
h_k =\sqrt{\alpha _k}h_0 + \sqrt{(1-\alpha _k)} \epsilon.
\end{align}
\end{small}

Hence, when a long decreasing sequence $\alpha _{1:K}$ is set such that $\alpha_K \approx 0$, the distribution of $h_K$ will converge to a standard Gaussian, i.e., $h_K \sim \mathcal{N}(\mathbf{0},\mathbf{I})$.
This indicates that the source signal $h_0$ will eventually be corrupted into Gaussian noise,
which conforms to the non-equilibrium thermodynamics phenomenon of the diffusion process \citeMain{sohl2015deep}.

Using the sample $h_0$ and noisy samples $\{h_k\}_{k=1}^K$ generated by forward diffusion, the diffusion model $g$ (which is often a deep network parameterized by $\theta$) is optimized to approximate the reverse diffusion process. 
Specifically, although the exact formulations may differ \citeMain{ho2020denoising,song2021denoising,sohl2015deep}, each reverse diffusion step can be expressed as a function $f$ that takes in $h_k$ and diffusion model $g$ as input to generate an output $h_{k-1}$ as follows:

\vspace{-0.2cm}
\begin{small}
\setlength{\abovedisplayskip}{4pt}
\setlength{\belowdisplayskip}{4pt}
\begin{align}
\label{eq: original reverse diffusion} 
h_{k-1} = f(h_k,g).
\end{align}
\end{small}

Finally, during testing, a Gaussian noise $h_K$ can be easily sampled, and the reverse diffusion step introduced in Eq.~\ref{eq: original reverse diffusion} can be recurrently performed to generate a high-quality sample $h_0$ using the trained diffusion model $g$.

\section{Proposed Method: DiffPose}
\label{sec:method}

\begin{figure*}[t]
  \centering
  \includegraphics[width=0.9\linewidth]{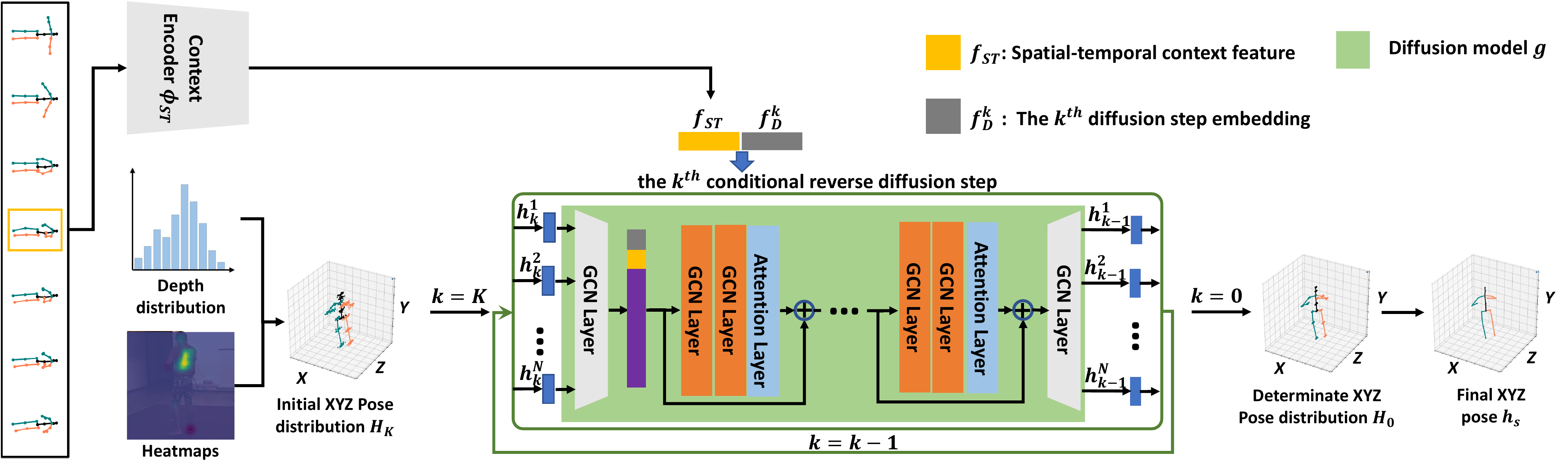}
  \vspace{-4mm}
  \caption{
  Illustration of our DiffPose framework during inference.
  {First, we use the Context Encoder $\phi_{ST}$ to extract the spatial-temporal context feature $f_{ST}$ from the given 2D pose sequence. 
  We also generate diffusion step embedding $f^k_D$ for each $k^{th}$ diffusion step.   
  Then, we initialize the indeterminate pose distribution $H_K$ using heatmaps derived from an off-the-shelf 2D pose detector and depth distributions that can either be computed from the training set or predicted by the Context Encoder $\phi_{ST}$.}
  Next, we sample $N$ noisy poses $\{h^i_K\}_{i=1}^N$ from $H_K$, which are required for performing distribution-to-distribution mapping.
  We feed these $N$ poses into the diffusion model $K$ times, where diffusion model $g$ is also conditioned on $f_{ST}$ and $f^k_D$ at each step, to obtain $\{h^i_0\}_{i=1}^N$ which represents the high-quality determinate distribution $H_0$. 
  Lastly, we use the mean of $\{h^i_0\}_{i=1}^N$ as our final 3D pose $h_s$.
  }
  \label{fig: Diffpose_architecture}
  \vspace{-6mm}
\end{figure*}

Given an RGB image frame $I_t$ or a video clip $V_t=\{I_\tau\}^{(t+T)}_{\tau = (t-T)}$, the goal of 3D human pose estimation is to predict the 3D coordinates of all the $J$ keypoints of the human body in $I_t$.
In this paper, inspired by diffusion-based generative models that can recurrently shed the indeterminacy in an initial distribution (e.g., Gaussian distribution) to reconstruct a high-quality determinate sample,  
we frame the 3D pose estimation task as constructing a determinate 3D pose distribution ($H_0$) from the highly indeterminate pose distribution ($H_K$) via diffusion models, which can handle the uncertainty and indeterminacy of 3D poses.

As shown in Fig. \ref{fig: Diffpose_architecture}, 
we conduct pose estimation in two stages:
(i) Initializing the indeterminate 3D pose distribution $H_K$ based on extracted heatmaps, which capture the underlying uncertainty of the input 2D pose in 3D space;
(ii) Performing the \textit{reverse diffusion process}, where we use a diffusion model $g$ to progressively denoise the initial distribution $H_K$ to a desired high-quality determinate distribution $H_0$, and then we can sample $h_0 \in \mathbb{R}^{3 \times J}$ from the pose distribution $H_0$ to synthesize the final 3D pose $h_s$.

In Sec. \ref{sec:initialize_hk}, we describe how to initialize the 3D distribution $H_K$ from an input 2D pose that effectively captures the uncertainty in the 3D space.
Then, we explain our forward diffusion process in Sec. \ref{subsec: human pose diffusion}
and the reverse diffusion process in Sec. \ref{subsec: human pose reverse diffusion}.
After that, we present the detailed training and testing process in Sec. \ref{subsec: diffpose inference}.
Finally, the architecture of our diffusion network is detailed in Sec. \ref{subsec: Diffusion Pose Network Architecture}.

\subsection{Initializing 3D Pose Distribution $H_K$}
\label{sec:initialize_hk}
In previous diffusion models \citeMain{ho2020denoising,song2021denoising,gu2022stochastic}, the reverse diffusion process often starts from random noise, which is progressively denoised to generate a high-quality output.
However, in 3D pose estimation, our input here is instead an estimated 2D pose that has its own uncertainty characteristics in 3D space.
To aid our diffusion model in handling the uncertainty and indeterminacy of each input 2D pose in 3D space,
we would like to initialize a corresponding 3D pose distribution $H_K$ that captures the uncertainty of the 3D pose.
Thus, the reverse diffusion process can start from the distribution $H_K$ with sample-specific knowledge (in contrast to Gaussian noise with no prior information), which leads to better performance.
Below, we describe how we construct the $x$, $y$, and $z$ uncertainty distribution for each joint of an input pose.

\textbf{Initializing $(x,y,z)$ distribution.} 
Intuitively, the $x$ and $y$ uncertainty distribution contains information regarding the likely regions in the image where the joints are located, and can roughly be seen as the outcome of ``outwards" diffusion from the ground-truth positions. 
It can be difficult to capture such 2D pose uncertainty distributions, which are often complicated and also vary for different joints of the given pose.
To address this, we take advantage of the available prior information to model the uncertainty of the 2D pose.
Notably, the 2D pose is often estimated from the image with an off-the-shelf 2D pose detector (e.g., CPN \citeMain{chen2018cascaded}), which first extracts heatmaps depicting the likely area on the image where each joint is located, before making predictions of 2D joint locations based on the extracted heatmaps.
Therefore, these heatmaps naturally reveal the uncertainty of the 2D pose predictions.
Hence, for the input 2D pose, we use the corresponding heatmaps from the off-the-shelf 2D pose detector as the $x$ and $y$ distribution. 

However, we are unable to obtain the $z$ distribution in the same way, as it is not known by the 2D pose detector.
Instead, one way we can compute the $z$ distribution is by calculating the occurrence frequencies of $z$ values in the training data, where we obtain a histogram for every joint.
We also explore another approach, where the uncertain $z$ distribution is initialized using the Context Encoder (which is introduced in Sec.~\ref{subsec: human pose reverse diffusion}), which we empirically observe to lead to faster convergence.

\subsection{Forward Pose Diffusion}
\label{subsec: human pose diffusion}
After initializing the indeterminate distribution $H_K$, the next step in our 3D pose estimation pipeline is to
progressively reduce the uncertainty $(H_K \rightarrow H_{K-1} \rightarrow ... \rightarrow H_0)$ using the diffusion model $g$ through the reverse diffusion process.
However, to attain the progressive denoising capability of the diffusion model $g$, we require ``ground truth" intermediate distributions as supervisory signals to train $g$.
Here, we obtain samples from these intermediate distributions via the \textit{forward diffusion process}, where we take a ground truth 3D pose distribution $H_0$ and gradually add noise to it, as shown in Fig. \ref{fig: Pose Reverse Diffusion}.
Specifically, given a desired determinate pose distribution $H_0$, we define the forward diffusion process as $(H_0 \rightarrow H_1 \rightarrow ... \rightarrow H_K)$, where $K$ is the maximum number of diffusion steps.
In this process, we aim to progressively increase the indeterminacy of $H_0$ towards the underlying pose uncertainty distribution $H_K$ as obtained in Sec. \ref{sec:initialize_hk}, 
such that we can obtain samples from intermediate distributions that correspond to $H_1,...,H_K$, which will allow us to optimize the diffusion model $g$ to smoothly perform the step-by-step denoising.

\textbf{DiffPose Forward Diffusion.}
For DiffPose, we do not want to diffuse our 3D pose towards a standard Gaussian noise. This is because our indeterminate distribution $H_K$ is not random noise, but is instead a $(x,y,z)$ distribution according to the 3D pose uncertainty, and has more complex characteristics.
This has several implications.
For example, the region of uncertainty for each joint and each coordinate of the initial pose distribution $H_K$ can be different.
Secondly, the mean locations of all joints should not be treated as equal to the origin (i.e., $0$ along all dimensions), due to the constraints of the body structure.
Due to these reasons, the basic generative diffusion process (in Sec.~\ref{sec:background}) cannot appropriately model the uncertainty of the initialized pose distribution $H_K$ (as described in Sec. \ref{sec:initialize_hk}) for our 3D pose estimation task, which motivates us to design a new forward diffusion process.

Designing such a forward diffusion process can be challenging, because the uncertainty distribution $H_K$, which is based on heatmaps, often has irregular and complex shapes, and it is not straightforward to express $H_K$ mathematically.
To overcome this, we propose to use a Gaussian Mixture Model (GMM) to model the uncertainty distribution $H_K$ for 3D pose estimation, as it can characterize intractable and complex distributions \citeMain{li1999mixture,nachmani2021non}, and is very effective to represent heatmap-based distributions \citeMain{wang2022low}.
Then, based on the fitted GMM model, we perform a corresponding GMM-based forward diffusion process.
Specifically, we set the number of Gaussian components in the GMM at $M$, and use the Expectation-Maximization (EM) algorithm to optimize the GMM parameters $\phi_{GMM}$ to fit the target distribution $H_K$ as follows: 

\begin{small}
\setlength{\abovedisplayskip}{4pt}
\setlength{\belowdisplayskip}{4pt}
\begin{align}
\max_{\phi_{GMM}} \,\,\,\,
&{\prod}_{i=1}^{N_{GMM}} {\sum}^M_{m=1}{\pi}_m \mathcal{N}_{pdf}(h^i_K | \mu_m, \Sigma_m),
\end{align}
\end{small} 
\noindent where $h^1_K,...,h^{N_{GMM}}_K$ are $N_{GMM}$ poses sampled from the pose distribution $H_K$,
and $\phi_{GMM} = \{\mu _1, \Sigma _1, \pi _1,...,\mu _M, \Sigma _M, \pi _M\}$ refers to the GMM parameters. 
Here, $\mu_m \in \mathbb{R}^{3J}$ and $\Sigma_m  \in \mathbb{R}^{3J \times 3J}$ are the mean values and covariance matrix of the $m^{th}$ Gaussian component. 
$\pi_m \in [0,1]$ is the probability that any sample $h^i_K$ is drawn from the $m^{th}$ mixture component ($\sum_{m=1}^M \pi_m = 1$).

Next, we want to run the forward diffusion process on the ground truth pose distribution $H_0$ such that after $K$ steps, the generated noisy distribution becomes equivalent to the fitted GMM distribution $\phi_{GMM}$, which we henceforth denote as $\hat{H}_K$ because it is a GMM-based representation of $H_K$.
To achieve this, we can modify Eq.~\ref{eq: original diffusion linear function} as follows:

\begin{small}
\setlength{\abovedisplayskip}{4pt}
\setlength{\belowdisplayskip}{4pt}
\begin{align}
\label{eq: modified diffusion linear function} 
\hat{h}_k = \mu^G + \sqrt{{\alpha}_k}(h_0- \mu^G) + \sqrt{(1-{\alpha_k})} \cdot \epsilon^G.
\end{align}
\end{small}
where $\hat{h}_k$ is a generated sample from the generated distribution $\hat{H}_k$ (which does not have a superscript since it describes how to generate a single sample), $\mu^G = \sum^M_{m=1} \textbf{1}_m \mu _m$, 
$\epsilon^G \sim \mathcal{N}(0, \sum^M_{m=1}(\textbf{1}_m \Sigma _m))$, and $\textbf{1}_m \in \{0,1\}$ is a binary indicator for the $m^{th}$ component such that $\sum^M_{m=1} \textbf{1}_m = 1$ and $Prob(\textbf{1}_m = 1) = \pi_m$.
In other words, we first select a component $\hat{m}$ via sampling according to the respective probabilities $\pi_m$, and set only $\textbf{1}_{\hat{m}}$ to 1.
Then, we sample the Gaussian noise from that component $\hat{m}$ using $\mu_{\hat{m}}$ and $\Sigma_{\hat{m}}$.
Notably, as $\alpha_K \approx 0$, $\hat{h}_K$ is drawn from the fitted GMM model, i.e., $\hat{h}_K = \mu^G + \epsilon^G \sim \mathcal{N}(\sum^M_{m=1}(\textbf{1}_m \mu_m), \sum^M_{m=1}(\textbf{1}_m \Sigma_m))$.
Thus, this allows us to generate samples from $\{\hat{H}_1,...,\hat{H}_K\}$ as supervisory signals.
More details can be found in Supplementary.

\subsection{Reverse Diffusion for 3D Pose Estimation}
\label{subsec: human pose reverse diffusion}
As shown in Fig.~\ref{fig: Pose Reverse Diffusion}, the reverse diffusion process aims to recover a determinate 3D pose distribution ${H}_0$ from the indeterminate pose distribution ${H}_K$, where $H_K$ has been discussed in Sec.~\ref{sec:initialize_hk}.
In the previous subsection, we represent ${H}_K$ via a GMM model to generate intermediate distributions $\{\hat{H}_1,...,\hat{H}_K\}$.
Here, we use these distributions to optimize our diffusion model $g$ (parameterized by $\theta$) to learn the reverse diffusion process 
$(\hat{H}_K \rightarrow ... \rightarrow \hat{H}_1 \rightarrow {H}_0)$, and progressively shed the indeterminacy from $\hat{H}_K$ to reconstruct the determinate source distribution $H_0$.
The architecture of the diffusion model $g$ is described in Sec.~\ref{subsec: Diffusion Pose Network Architecture}.

\textbf{Context Encoder $\phi_{ST}$.}
However, it is difficult to directly perform the reverse diffusion process 
using only $\hat{H}_K$ as the input of the diffusion model $g$. 
This is because $g$ will not observe much context information from the input videos/images, leading to difficulties for $g$ to generate accurate poses from the indeterminate distribution $H_K$. 
Therefore, we propose to utilize the available context information from the input to guide $g$ to achieve more accurate predictions. 
The context information can constrain the model's denoising based on the observed inputs, and guide the model to produce more accurate predictions.

Specifically, to guide the diffusion model $g$, we leverage the \textit{spatial-temporal context}.
The context information can be extracted from the 2D pose sequence derived from $V_t$ (or just a single 2D pose derived from $I_t$ if $V_t$ is not available).
This context information aids the reverse diffusion process, providing additional information to the diffusion model $g$ that helps to reduce uncertainty and generate more accurate 3D poses.
To achieve that, we introduce the Context Encoder $\phi_{ST}$ to extract spatial-temporal information $f_{ST}$ from the 2D pose sequence, and condition the reverse diffusion process on $f_{ST}$ (as shown in Fig.~\ref{fig: Diffpose_architecture}).

\textbf{Reverse Diffusion Process.}
Overall, our reverse diffusion process aims to recover a determinate pose distribution $H_0$ from the indeterminate pose distribution $\hat{H}_K$ (during training) or $H_K$ (during testing).
Here, we describe the reverse diffusion process during training and use $\hat{H}_K$ notation.
We first use Context Encoder $\phi_{ST}$ to extract $f_{ST}$ from the 2D pose sequence.
Moreover, to allow the diffusion model to learn to denoise samples appropriately at each diffusion step, we also generate the unique step embedding $f^k_D$ to represent the $k^{th}$ diffusion step via the sinusoidal function. 
Then, for a noisy pose $\hat{h}_k$ sampled from $\hat{H}_k$, we use diffusion model $g$,
conditioned on the diffusion step $k$ and the spatial-temporal context feature $f_{ST}$,
to progressively reconstruct $\hat{h}_{k-1}$ from $\hat{h}_k$ as follows:
\begin{small} 
\begin{align}
\label{eq: sample} 
\hat{h}_{k-1} = &  g_\theta (\hat{h}_{k},f_{ST},f^k_D),~~k \in \{1,...,K \}.
\end{align}
\end{small}

\subsection{Overall Training and Testing Process}
\label{subsec: diffpose inference}
Overall, for each sample during training, we (i) initialize $H_K$; (ii) use $H_0$ and $H_K$ to generate supervisory signals $\{\hat{H}_1,...,\hat{H}_K \}$ via the forward process; (iii) run $K$ steps of the reverse process starting from $\hat{H}_K$ and optimize with our generated signals.
During testing, we (i) initialize $H_K$; (ii) run $K$ steps of the reverse process starting from $H_K$ to obtain final prediction $h_s$. More details are described below.

\textbf{Training.}
First, from the input sequence $V_t$ (or frame $I_t$), we extract the 2D heatmaps together with the estimated 2D pose via an off-the-shelf 2D pose detector \citeMain{chen2018cascaded}.
Then, we compute the $z$ distribution, either from the training set or predicted by the Context Encoder $\phi_{ST}$.
After that, we initialize $H_K$ based on the 3D distribution for each joint and use the EM algorithm to get the best-fit GMM parameters $\phi_{GMM}  = \{\mu _1, \Sigma _1, \pi _1,...,\mu _M, \Sigma _M, \pi _M\}$ for $H_K$.
Based on $\phi_{GMM}$, we use the ground truth 3D pose $h_0$ to \textit{directly} generate $N$ sets of $\hat{h}_1,...,\hat{h}_K$ via Eq.~\ref{eq: modified diffusion linear function}, i.e., $\big\{\{\hat{h}_1^i,...,\hat{h}_K^i \} \big\}_{i=1}^N$. 
Specifically, we first sample a component $\hat{m}^i$ for each $i^{th}$ set according to probabilities $\{\pi_m \}_{m=1}^M$, and use the $\hat{m}^i$-th Gaussian component to directly add noise for the $i^{th}$ set $\{\hat{h}_1^i,...,\hat{h}_K^i \}$. 
Next, we extract the spatial-temporal context $f_{ST}$ using the Context Encoder $\phi_{ST}$.
Then, we want to optimize the model parameters $\theta$ to reconstruct $\hat{h}_{k-1}^i$ from $\hat{h}_{k}^i$ in a step-wise manner. Following previous works on diffusion models \citeMain{ho2020denoising,song2021denoising}, we formulate our loss $\mathcal{L}$ as follows (where $\hat{h}_0^i = h_0$ for all $i$):

\vspace{-3mm}
\begin{small}
\setlength{\abovedisplayskip}{1pt}
\setlength{\belowdisplayskip}{1pt}
\label{eq: objective function}
\begin{align}
\mathcal{L}  =  \sum_{i=1}^N \sum_{k=1}^{K} 
\norm{g_\theta (\hat{h}_{k}^i,f_{ST},f^k_D) - \hat{h}_{k-1}^i}_2^2 .
\end{align}
\end{small}

\textbf{Testing.}
Similar to the start of the training procedure, during testing we first initialize $H_K$ and also extract $f_{ST}$.
Then, we perform the reverse diffusion process, where we sample $N$ poses from $H_K$ ($h^1_K,h^2_K,...,h^N_K$) and recurrently feed them into diffusion model $g$ for $K$ times, to obtain $N$ high-quality 3D poses ($h^1_0,h^2_0,...,h^N_0$).
We need $N$ noisy poses here, because we are mapping from a distribution to another distribution.
Then, to obtain the final high-quality and reliable pose $h_s$, we calculate the mean of the $N$ denoised samples $\{h^1_0, \dots, h^N_0\}$.

\subsection{DiffPose Architecture}
\label{subsec: Diffusion Pose Network Architecture}
Our framework consists of two sub-networks: a diffusion network $(g)$ that performs the steps in the reverse process and a Context Encoder $(\phi_{ST})$ that extracts the context feature from the 2D pose sequence (or frame).

\textbf{Main Diffusion Model $g$.}
We adopt a lightweight GCN-based architecture for $g$ to perform 3D pose estimation via diffusion, which is modified from \citeMain{zhao2022graformer}.
The graph convolution layer treats the human skeleton as a graph (with joints as the nodes), and effectively encodes topological information between joints for 3D human pose estimation.
Moreover, we interlace GCN layers with Self-Attention layers, which can encode global relationships between non-adjacent joints and allow for better structural understanding of the 3D human pose as a whole.
As shown in Fig.~\ref{fig: Diffpose_architecture}, our diffusion model $g$ mainly consists of 3 stacked GCN-Attention Blocks with residual connections, where each GCN-Attention Block comprises of two standard GCN layers and a Self-Attention layer. A GCN layer is added at the front and back of these stacked GCN-Attention Blocks to control the embedding size of GCN-Attention Blocks.

Specifically, the starting GCN layer maps the input $h_k \in \mathbb{R}^{J\times 3}$ to a latent embedding $E \in \mathbb{R}^{J \times 128}$.
On the other hand, we extract spatial-temporal context information $f_{ST} \in \mathbb{R}^{J \times 128}$.
In order to provide information to the model regarding the current step number $k$, we also generate a diffusion step embedding $f_D^k \in \mathbb{R}^{J \times 256}$ using the sinusoidal function.
Then, we combine these embeddings to form features $v_1 \in \mathbb{R}^{J \times 256}$, where $E$ and $f_{ST}$ are first concatenated along the second dimension, before adding $f_D^k$ to the result.
Features $v_1$ are then fed into the stack of 3 GCN-Attention Blocks, which all have the exact same structure. 
The output features from the last GCN-Attention Block are fed into the final GCN layer to be mapped into an output pose $h_{k-1} \in \mathbb{R}^{J\times 3}$. 
Then, we feed $h_{k-1}$ back to $g$ as input again to perform another reverse step. At the final $K$-th step, we obtain an output pose $h_{0} \in \mathbb{R}^{J\times 3}$.

\textbf{Context Encoder $\phi_{ST}$.}
In this paper, we leverage a transformer-based network \citeMain{zhang2022mixste} to capture the spatial-temporal context information in the 2D pose sequence $V_t$. 
Note that, if we do not have the video, we only input a single frame $I_t$, and utilize \citeMain{zhao2022graformer} instead.

\section{Experiments}
We evaluate our method on two widely used datasets for 3D human pose estimation: Human3.6M~\citeMain{ionescu2013human3} and MPI-INF-3DHP~\citeMain{mehta2017monocular}. Specifically, we conduct experiments to evaluate the performance of our method in two scenarios: video-based and frame-based 3D pose estimation.

\textbf{Human3.6M} \citeMain{ionescu2013human3} is the largest benchmark for 3D human pose estimation, consisting of 3.6 million images captured from four cameras, where 15 daily activities are performed by 11 subjects. 
For video-based 3D pose estimation, we follow previous works~\citeMain{pavllo20193d,liu2020attention,chen2021anatomy} to train on five subjects (S1, S5, S6, S7, S8) and test on two subjects (S9 and S11). For frame-based 3D pose estimation, we follow ~\citeMain{zhao2019semantic,zhao2022graformer,xu2021graph} to train on (S1, S5, S6, S7, S8) subjects and test on (S9, S11) subjects.
We report the mean per joint position error (MPJPE) and Procrustes MPJPE (P-MPJPE). The former computes the Euclidean distance between the predicted joint positions and the ground truth positions. The latter is the MPJPE after the predicted results are aligned to the ground truth via a rigid transformation. Due to page limitations, we move P-MPJPE results to Supplementary.

\textbf{MPI-INF-3DHP} \citeMain{mehta2017monocular} is a large 3D pose dataset captured in both indoor and outdoor environments, with 1.3 million frames. 
Following \citeMain{mehta2017monocular,lin2019trajectory,chen2021anatomy,zheng20213d}, we train DiffPose using all activities from 8 camera views in the training set and evaluate on valid frames in the test set.
Here, we report metrics of MPJPE, Percentage of Correct Keypoints (PCK) with the threshold of 150 $mm$, and Area Under Curve (AUC) for a range of PCK thresholds to compare our performance with other methods on the video-based setting.

\textbf{Implementation Details.}
We set the number of pose samples $N$ to 5 and number of reverse diffusion steps $K$ to 50. We fit $\hat{H}_K$ via a GMM model with 5 kernels ($M=5$) for forward diffusion, and accelerate our diffusion inference procedure for all experiments via an acceleration technique DDIM \citeMain{song2021denoising}, where only five steps are required to complete the reverse diffusion process.
For video pose estimation, we set the Context Encoder $\phi_{ST}$ to follow \citeMain{zhang2022mixste}, and for frame-based pose estimation, we set $\phi_{ST}$ to follow \citeMain{zhao2022graformer}. 
The Context Encoder $\phi_{ST}$ is pre-trained on the training set 
to predict $(x,y,z)$, then frozen during diffusion model training; we use it to produce features $f_{ST}$ and also to initialize the $z$ distribution.
For video-based pose estimation, we follow ~\citeMain{pavllo20193d,cai2019exploiting} to use detected 2D pose (using CPN~\citeMain{chen2018cascaded}) and ground truth 2D pose on Human3.6M, and use ground truth 2D pose on MPI-INF-3DHP.
For frame-based pose estimation, we follow ~\citeMain{zhao2019semantic,zhao2022graformer} to use the 2D pose detected by \citeMain{chen2018cascaded} and ground truth 2D pose to conduct experiments on Human3.6M.
More details are in Supplementary.

\subsection{Comparison with State-of-the-art Methods}

\textbf{Video-based Results on Human3.6M.}
We follow \citeMain{pavllo20193d,zhang2022mixste,zeng2020srnet} to use 243 frames for 3D pose estimation and compare our method against existing works on Human3.6M in Tab. \ref{tab: Human3.6M video}.
As shown in the top of Tab. \ref{tab: Human3.6M video}, our method achieves the best MPJPE results using the detected 2D pose, and significantly outperforms the SOTA method \citeMain{zhang2022mixste} by around 4 $mm$.
This shows that DiffPose can effectively improve monocular 3D pose estimation.
Moreover, we also conduct experiments using the ground truth 2D pose as input, and report our results at the bottom of Tab.~\ref{tab: Human3.6M video}.
Our DiffPose again outperforms all previous methods by a large margin.

\begin{table}[t]
\normalsize
\centering
\tabcolsep=0.3mm
\caption
{
Video-based results on Human3.6M in millimeters under MPJPE. Top table shows the results on detected 2D poses. Bottom table shows the results on ground truth 2D poses.
} 
\label{tab: Human3.6M video}
\vspace{-3mm}
\resizebox{0.483\textwidth}{!}{
\begin{tabular}{@{}l|ccccccccccccccc|c@{}}
\hline
\noalign{\smallskip}
MPJPE(CPN) & Dir & Disc & Eat & Greet & Phone & Photo & Pose & Pur & Sit & SitD & Smoke & Wait & WalkD & Walk & WalkT & Avg \\
\noalign{\smallskip}
\hline
\noalign{\smallskip}
Pavllo \citeMain{pavllo20193d} &45.2&46.7&43.3&45.6&48.1&55.1&44.6&44.3&57.3&65.8&47.1&44.0&49.0&32.8&33.9&46.8 \\
Liu \citeMain{liu2020attention} &41.8&44.8&41.1&44.9&47.4&54.1&43.4&42.2&56.2&63.6&45.3&43.5&45.3&31.3&32.2&45.1 \\
Zeng \citeMain{zeng2020srnet}
& 46.6&47.1&43.9&41.6&45.8&{49.6}&46.5&40.0&53.4&61.1&46.1&42.6&43.1&31.5&32.6&44.8 \\
Zheng \citeMain{zheng20213d}
&41.5&44.8&39.8&42.5&46.5&51.6&42.1&42.0&{53.3}&60.7&45.5&43.3&46.1&31.8&32.2&44.3\\
Li \citeMain{li2022mhformer}
& 39.2 & 43.1 & 40.1 & 40.9 & 44.9 & 51.2 & 40.6 & 41.3 & 53.5 & 60.3 & 43.7 & 41.1 & 43.8 & 29.8 & 30.6 & 43.0\\
Shan \citeMain{shan2022p} 
&38.4&{42.1}&39.8&{40.2}&{45.2}&\underline{48.9}&{40.4}&\underline{38.3}&53.8&{57.3}&{43.9}&{41.6}&{42.2}&{29.3}&{29.3}&{42.1}\\
Zhang \citeMain{zhang2022mixste}
& \underline{37.6} & \underline{40.9} & \underline{37.3} & \underline{39.7} & \underline{42.3} & {49.9} & \underline{40.1} & {39.8} & \underline{51.7} & \underline{55.0} & \underline{42.1} & \underline{39.8} & \underline{41.0} & \underline{27.9} & \underline{27.9} & \underline{40.9}\\
\noalign{\smallskip}
\hline
\noalign{\smallskip}
Ours & \textbf{33.2} &\textbf{36.6}& \textbf{33.0}&\textbf{35.6}&\textbf{37.6}&\textbf{45.1}&\textbf{35.7}&\textbf{35.5}&\textbf{46.4}&\textbf{49.9}&\textbf{37.3}&\textbf{35.6}&\textbf{36.5}&\textbf{24.4}&\textbf{24.1}&\textbf{36.9}\\

\noalign{\smallskip}
\hline
\hline
\noalign{\smallskip}
MPJPE(GT) & Dir & Disc & Eat & Greet & Phone & Photo & Pose & Pur & Sit & SitD & Smoke & Wait & WalkD & Walk & WalkT & Avg \\
\noalign{\smallskip}
\hline
\noalign{\smallskip}
Pavllo \citeMain{pavllo20193d} &35.2&40.2&32.7&35.7&38.2&45.5&40.6&36.1&48.8&47.3&37.8&39.7&38.7&27.8&29.5&37.8  \\
Liu \citeMain{liu2020attention} &34.5&37.1&33.6&34.2&32.9&37.1&39.6&35.8&40.7&41.4&33.0&33.8&33.0&26.6&26.9&34.7   \\
Zeng \citeMain{zeng2020srnet} 
&34.8&32.1&{28.5}&30.7&31.4&36.9&35.6&{30.5}&38.9&40.5&32.5&31.0&29.9&{22.5}&24.5&32.0   \\
Zheng \citeMain{zheng20213d}
&30.0&33.6&29.9&31.0&{30.2}&{33.3}&34.8&31.4&37.8&38.6&31.7&31.5&29.0&23.3&{23.1}&31.3\\
Li \citeMain{li2022mhformer} 
& 27.7 & 32.1 & 29.1 & 28.9 & 30.0 & 33.9 & 33.0 & 31.2 & 37.0 & 39.3 & 30.0 & 31.0 & 29.4 & 22.2 & 23.0 & 30.5\\
Shan \citeMain{shan2022p}
&{28.5}&{30.1}&{28.6}&{27.9}&{29.8}&{33.2}&{31.3}&{27.8}&{36.0}&{37.4}&{29.7}&{29.5}&{28.1}&{21.0}&{21.0}&{29.3}\\
Zhang \citeMain{zhang2022mixste}
& \underline{21.6} & \underline{22.0} & \underline{20.4} & \underline{21.0} & \underline{20.8} & \underline{24.3} & \underline{24.7} & \underline{21.9} & \underline{26.9} & \underline{24.9} & \underline{21.2} & \underline{21.5} & \underline{20.8} & \underline{14.7} & \underline{15.7} & \underline{21.6}\\
\noalign{\smallskip}
\hline
\noalign{\smallskip}
Ours &\textbf{18.6}&\textbf{19.3}&\textbf{18.0}&\textbf{18.4}&\textbf{18.3}&\textbf{21.5}&\textbf{21.5}&\textbf{19.1}&\textbf{23.6}&\textbf{22.3}&\textbf{18.6}&\textbf{18.8}&\textbf{18.3}&\textbf{12.8}&\textbf{13.9}&\textbf{18.9}\\
\hline
\noalign{\smallskip}
\end{tabular}
}
\vspace{-3mm}
\end{table}
\setlength{\intextsep}{0pt}%
\setlength{\columnsep}{6pt}%
\begin{wraptable}{r}{0.21\textwidth}
\footnotesize
\caption
{Video-based results on MPI-INF-3DHP.}
\vspace{-3mm}
\label{table: MPII-3DHP video}
\tabcolsep=0.3mm
\resizebox{\linewidth}{!}{
\begin{tabular}{@{}l|ccc@{}}
\hline
Method & PCK $\uparrow$ & AUC $\uparrow$ & MPJPE $\downarrow$\\
\noalign{\smallskip}
\hline
\noalign{\smallskip}
Pavllo \citeMain{pavllo20193d}
&86.0&51.9&84.0\\
Wang \citeMain{wang2020motion} 
&86.9&62.1&68.1\\
Zheng \citeMain{zheng20213d} 
&{88.6}&56.4&77.1\\
Li \citeMain{liu2020attention} 
& 93.8 & 63.3 & 58.0 \\
Zhang \citeMain{zhang2022mixste} 
& \underline{94.4} & \underline{66.5 }& \underline{54.9}  \\
\noalign{\smallskip}
\hline
\noalign{\smallskip}
Ours 
&\textbf{98.0}&\textbf{75.9}&\textbf{29.1}\\
\hline
\end{tabular}
}%
\end{wraptable}
\textbf{Video-based Results on MPI-INF-3DHP.}
We also evaluate our method on MPI-INF-3DHP. Here, we use 81 frames as our input due to the shorter video length of this dataset.
The results in Tab.~\ref{table: MPII-3DHP video} demonstrate that our method achieves the best performance, showing the efficacy of our DiffPose in improving performance in outdoor scenes.

\begin{table}[t]
\normalsize
\centering
\tabcolsep=0.3mm
\caption
{
Frame-based results on Human3.6M in millimeters under MPJPE. Top table shows the results on detected 2D poses. Bottom table shows the results on ground truth 2D poses.
} 
\vspace{-3mm}
\label{tab: Human3.6M Frame}
\resizebox{0.483\textwidth}{!}{
\begin{tabular}{@{}l|ccccccccccccccc|c@{}}
\hline
\noalign{\smallskip}
MPJPE(CPN) & Dir & Disc & Eat & Greet & Phone & Photo & Pose & Pur & Sit & SitD & Smoke & Wait & WalkD & Walk & WalkT & Avg \\
\noalign{\smallskip}
\hline
\noalign{\smallskip}
Pavlakos \citeMain{pavlakos2017coarse}
& 67.4 & 71.9 & 66.7 & 69.1 & 72.0 & 77.0 & 65.0 & 68.3 & 83.7 & 96.5 & 71.7 & 65.8 & 74.9 & 59.1 & 63.2 & 71.9\\
Martinez\citeMain{martinez2017simple}
& 51.8 & 56.2 & 58.1 & 59.0 & 69.5 & 78.4 & 55.2 & 58.1 & 74.0 & 94.6 & 62.3 & 59.1 & 65.1 & 49.5 & 52.4 & 62.9\\
Sun \citeMain{sun2017compositional}
& 52.8 & 54.8 & 54.2 & 54.3 & 61.8 & \textbf{53.1} & 53.6 & 71.7 & 86.7 & \underline{61.5} & 67.2 & 53.4 & \underline{47.1} & 61.6 & 53.4 & 59.1\\
Yang \citeMain{yang20183d}
& 51.5 & 58.9 & 50.4 & 57.0 & 62.1 & 65.4 & 49.8 & 52.7 & 69.2 &     85.2 & 57.4 & 58.4 & \textbf{43.6} & 60.1 & 47.7 & 58.6\\
Hossain \citeMain{hossain2018exploiting}
& 48.4 & 50.7 & 57.2 & 55.2 & 63.1 & 72.6 & 53.0 & 51.7 & 66.1 & 80.9 & 59.0 & 57.3 & 62.4 & 46.6 & 49.6 & 58.3\\
Zhao \citeMain{zhao2019semantic}
& 48.2 & 60.8 & 51.8 & 64.0 & 64.6 & \underline{53.6} & 51.1 & 67.4 & 88.7 & \textbf{57.7} & 73.2 & 65.6 & 48.9 & 64.8 & 51.9 & 60.8\\
Liu \citeMain{liu2020comprehensive}
& 46.3 & 52.2 & \underline{47.3} & 50.7 & 55.5 & 67.1 & 49.2 & \underline{46.0} & 60.4 & 71.1 & 51.5 & 50.1 & 54.5 & 40.3 & 43.7 & 52.4 \\
Xu \citeMain{xu2021graph}
& \underline{45.2} & \underline{49.9} & 47.5 & 50.9 & 54.9 & 66.1 & 48.5 & 46.3 & \underline{59.7} & 71.5 & \underline{51.4} & \underline{48.6} & 53.9 & 39.9 & 44.1 & 51.9\\
Zhao \citeMain{zhao2022graformer}
& \underline{45.2} & 50.8 & 48.0 & \underline{50.0} & \underline{54.9} & 65.0 & \underline{48.2} & 47.1 & 60.2 & 70.0 & 51.6 & 48.7 & 54.1 & \underline{39.7} & \underline{43.1} & \underline{51.8}\\
\noalign{\smallskip}
\hline
\noalign{\smallskip}
Ours    & \textbf{42.8} & \textbf{49.1} & \textbf{45.2} & \textbf{48.7} & \textbf{52.1} & {63.5} & \textbf{46.3} & \textbf{45.2} & \textbf{58.6} & 66.3 & \textbf{50.4} & \textbf{47.6} & 52.0 & \textbf{37.6} & \textbf{40.2} & \textbf{49.7}\\
\noalign{\smallskip}
\hline
\hline
\noalign{\smallskip}
MPJPE(GT) & Dir & Disc & Eat & Greet & Phone & Photo & Pose & Pur & Sit & SitD & Smoke & Wait & WalkD & Walk & WalkT & Avg \\
\noalign{\smallskip}
\hline
\noalign{\smallskip}
Martinez \citeMain{martinez2017simple}
& 37.7 & 44.4 & 40.3 & 42.1 & 48.2 & 54.9 & 44.4 & 42.1 & 54.6 & 58.0 & 45.1 & 46.4 & 47.6 & 36.4 & 40.4 & 45.5 \\
Hossain \citeMain{hossain2018exploiting} 
& 35.2 & 40.8 & 37.2 & 37.4 & 43.2 & 44.0 & 38.9 & 35.6 & 42.3 & 44.6 & 39.7 & 39.7 & 40.2 & 32.8 & 35.5 & 39.2 \\
Zhao \citeMain{zhao2019semantic}
& 37.8 & 49.4 & 37.6 & 40.9 & 45.1 & \underline{41.4} & 40.1 & 48.3 & 50.1 & \underline{42.2} & 53.5 & 44.3 & 40.5 & 47.3 & 39.0 & 43.8 \\
Liu \citeMain{liu2020comprehensive}
& 36.8 & 40.3 & 33.0 & 36.3 & 37.5 & 45.0 & 39.7 & 34.9 & 40.3 & 47.7 & 37.4 & 38.5 & 38.6 & 29.6 & 32.0 & 37.8 \\
Xu \citeMain{xu2021graph}
&  35.8 & 38.1 & 31.0 & 35.3 & 35.8 & 43.2 & 37.3 & 31.7 & 38.4 & 45.5 & 35.4 & 36.7 & 36.8 & 27.9 & 30.7 & 35.8\\
Zhao \citeMain{zhao2022graformer}
& \underline{32.0} & \underline{38.0} & \underline{30.4} & \underline{34.4} & \underline{34.7} & 43.3 & \underline{35.2} & \underline{31.4} & \underline{38.0} & 46.2 & \underline{34.2} & \underline{35.7} & \underline{36.1} & \underline{27.4} & \underline{30.6} & \underline{35.2} \\
\noalign{\smallskip}
\hline
\noalign{\smallskip}
Ours  &\textbf{28.8}&\textbf{32.7}&\textbf{27.8}&\textbf{30.9}&\textbf{32.8}&\textbf{38.9}&\textbf{32.2}&\textbf{28.3}&\textbf{33.3}&\textbf{41.0}&\textbf{31.0}&\textbf{32.1}&\textbf{31.5}&\textbf{25.9}&\textbf{27.5}&\textbf{31.6}\\
\hline
\noalign{\smallskip}
\end{tabular}
\vspace{-100mm}
}
\end{table}

\textbf{Frame-based Results on Human3.6M.}
To further investigate the efficacy of DiffPose, we evaluate it in a more challenging setting: frame-based 3D pose estimation.
Here, we only extract context information from the single input frame via our Context Encoder $\phi_{ST}$.
Our results on Human3.6M are reported in Tab.~\ref{tab: Human3.6M Frame}.
As shown at the top of Tab.~\ref{tab: Human3.6M Frame}, our DiffPose surpasses all existing methods in average MPJPE using detected 2D poses. 
At the bottom of Tab. \ref{tab: Human3.6M Frame}, we observe that DiffPose also outperforms all methods with a large margin when ground truth 2D poses are used.

\begin{figure}[t]
  \centering
  \includegraphics[width=1\linewidth]{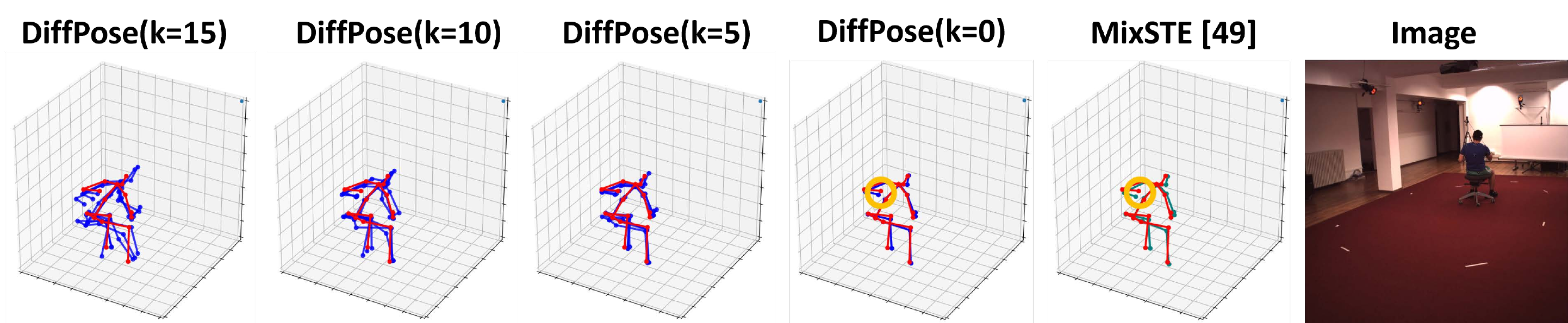}
  \vspace{-4mm}
  \caption{Qualitative results. Red colored 3D pose corresponds to the ground truth. Under occlusion, our DiffPose predicts a pose that is more accurate than previous methods (circled in orange).
  }
  \vspace{-4mm}
  \label{fig: Quantative resutls}
\end{figure}

\textbf{Qualitative results.} 
In the first four columns of Fig.~\ref{fig: Quantative resutls}, we provide visualizations of the reverse diffusion process, where the step $k$ decreases from 15 to 0. 
We observe that DiffPose can progressively narrow down the gap between the sampled poses and the ground-truth pose. Moreover, we compare our method with the current SOTA method \citeMain{zhang2022mixste}, which shows that our method can generate more reliable 3D pose solutions, especially for ambiguous body parts.

\subsection{Ablation Study}
To verify the impact of each proposed design, we conduct extensive ablation experiments on Human3.6M dataset using the detected 2D poses in the video-based setting.

\setlength{\intextsep}{0pt}%
\setlength{\columnsep}{6pt}%
\begin{wraptable}{r}{0.20\textwidth}
\caption{Ablation study for diffusion pipeline.}
\label{table: Ablation Diffusion}
\vspace{-3mm}
\resizebox{\linewidth}{!}{%
\begin{tabular}{@{}l|cc@{}}
\hline
Method & MPJPE & P-MPJPE\\
\hline
Baseline A & 44.3 & 33.7\\
Baseline B & 41.1 & 32.8\\
\hline
DiffPose & 36.9 & 28.6\\
\hline
\end{tabular}
}%
\end{wraptable}
\textbf{Impact of Diffusion Process.} We first evaluate the diffusion process's effectiveness. Here we build two baseline models: 
(1) \textbf{Baseline A}: It has the same structure as our diffusion model but the 3D pose estimation is conducted in a single step.
(2) \textbf{Baseline B}: It has the nearly same architecture as our diffusion model but the diffusion model is stacked multiple times to approximate the computational complexity of DiffPose. {Note that both baselines are optimized to predict 3D human pose instead of learning the reverse diffusion process.}
We report the results of the baselines and DiffPose in Tab.~\ref{table: Ablation Diffusion}. 
The performance of both baselines are much worse than our DiffPose, which indicates that the performance improvement of our method comes from the designed diffusion pipeline.

\setlength{\intextsep}{0pt}%
\setlength{\columnsep}{6pt}%
\begin{wraptable}{r}{0.21\textwidth}
\caption{Ablation study for GMM design}
\label{table: Ablation GMM}
\vspace{-3mm}
\resizebox{\linewidth}{!}{%
\begin{tabular}{@{}l|cc@{}}
\hline
Method & MPJPE & P-MPJPE\\
\hline
Stand-Diff & 40.1 & 31.1\\
GMM-Diff(M=1) & 38.0 & 29.7\\
GMM-Diff(M=5) & 36.9 & 28.6\\
GMM-Diff(M=9) & 36.5 & 28.5\\
\hline
\end{tabular}
}%
\end{wraptable}
\textbf{Impact of GMM.}
To validate the effect of the GMM design, we consider two alternative ways to train our diffusion model: 
(1) \textbf{Stand-Diff}: we directly adopt the basic forward diffusion process introduced in Eq.~\ref{eq: original diffusion linear function} for model training. 
(2) \textbf{GMM-Diff}: we utilize GMM to fit the initial 3D pose distribution $H_K$ to generate intermediate distributions for model training. Moreover, we test the number of kernels in GMM $M$ (from 1 to 9) to investigate the characteristics of GMM in pose diffusion.
We report the results with different $M$ in Tab.~\ref{table: Ablation GMM}. 
Experiments show that our GMM-based design significantly outperforms the baseline Stand-Diff, which shows the effectiveness of using a GMM to approximate $H_K$.
Moreover, we can observe that using 5 kernels ($M=5$) is sufficient to effectively capture the uncertainty distribution.

\setlength{\intextsep}{0pt}%
\setlength{\columnsep}{6pt}%
\begin{wraptable}{r}{0.24\textwidth}
\caption{
Ablation study for $f_{ST}$.
}
\label{table: Ablation condition}
\vspace{-3mm}
\resizebox{\linewidth}{!}{%
\begin{tabular}{@{}l|cc@{}}
\hline
Method & MPJPE & P-MPJPE\\
\hline
\citeMain{shan2022p} & 42.1 & 34.4\\
Ours + \citeMain{shan2022p} & 39.3 & 31.8\\
\hline
\citeMain{zhang2022mixste} & 40.9 & 32.6\\
Ours + \citeMain{zhang2022mixste} & 36.9 & 28.7\\
\hline
\end{tabular}
}%
\end{wraptable}
\textbf{Impact of context $f_{ST}$.}
Another crucial component to explore is the role of spatial-temporal context $f_{ST}$ in our method.
Here, we evaluate the performance when using various context encoders \citeMain{shan2022p, zhang2022mixste} to obtain $f_{ST}$. 
As shown in Tab.~\ref{table: Ablation condition}, our DiffPose achieves good performance using both models.
We also find that DiffPose significantly outperforms both context encoders, which verifies the efficacy of our approach.

\setlength{\intextsep}{0pt}%
\setlength{\columnsep}{6pt}%
\begin{wrapfigure}{r}{0.17\textwidth}
\includegraphics[width=0.17\textwidth]{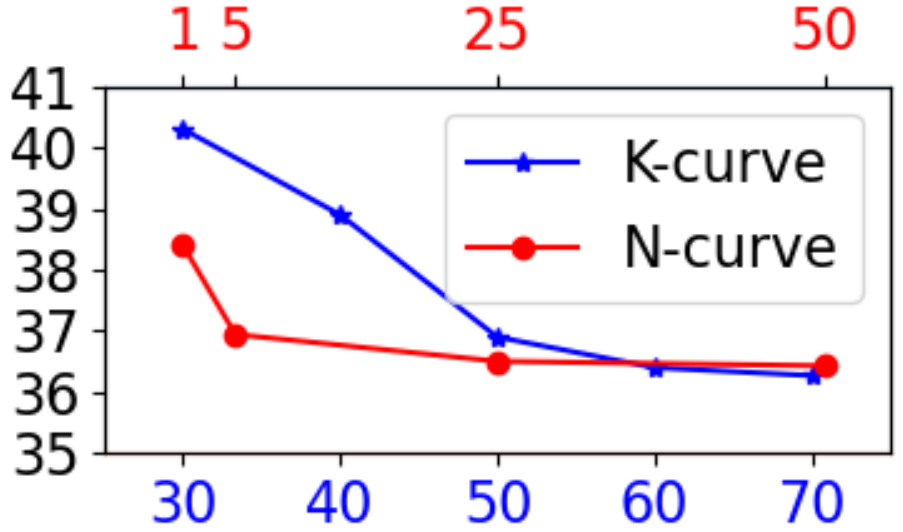}
\vspace{-5mm}
\caption{Evaluation of parameters $K$ and $N$.}
\label{figure: ablation k and n}
\end{wrapfigure}

\textbf{Impact of reverse diffusion steps $K$ and sample number $N$.}
To further investigate the characteristics of our pose diffusion process, we conduct several experiments with different diffusion step numbers ($K$) and sample numbers ($N$) and plot the results in Fig.~\ref{figure: ablation k and n}.
We observe that MPJPE first drops significantly till $K=50$, and shows minor improvements when $K>50$.
Thus, we use 50 diffusion steps ($K=50$) in our method, which can effectively and efficiently shed indeterminacy.
On the other hand, we find that model performance improves with the number of samples $N$ until $N=5$, where our performance stays roughly consistent.

\setlength{\intextsep}{0pt}%
\setlength{\columnsep}{6pt}%
\begin{wraptable}{r}{0.21\textwidth}
\caption{
Analysis of speed. Our method can run efficiently, yet outperforms SOTA significantly.
}
\label{table: Ablation speed}
\vspace{-3mm}
\resizebox{\linewidth}{!}{%
\begin{tabular}{@{}l|cc@{}}
\hline
Method & MPJPE & FPS\\
\hline
Li \citeMain{li2022mhformer} & 43.0 & 328 \\
Zhang \citeMain{zhang2022mixste} & 40.9 & 974\\
\hline
{DiffPose w/o DDIM} & 36.7 & 173 \\
{DiffPose w/ DDIM} & 36.9 & 671 \\
\hline
\end{tabular}
}%
\end{wraptable}

\textbf{Inference Speed.}
In Tab.~\ref{table: Ablation speed}, we compare the speed of DiffPose with existing methods in terms of Frames Per Second (FPS). 
Our DiffPose with DDIM acceleration can achieve a competitive speed compared with the current SOTA \citeMain{zhang2022mixste} while obtaining better performance. Moreover, even without DDIM acceleration, the FPS of our model is still higher than 170 FPS, which satisfies most real-time requirements.

\section{Conclusion}
This paper presents DiffPose, a novel diffusion-based framework that handles the uncertainty and indeterminacy in monocular 3D pose estimation. 
DiffPose first initializes the indeterminate 3D pose distribution and then recurrently sheds the indeterminacy in this distribution to obtain the final high-quality 3D human pose distribution for reliable pose estimation.
Extensive experiments show that the proposed DiffPose achieves state-of-the-art performance on two widely used benchmark datasets.

\noindent
\footnotesize
\textbf{Acknowledgments.}
This work is supported by MOE AcRF Tier 2 (Proposal ID: T2EP20222-0035), National Research Foundation Singapore under its AI Singapore Programme (AISG-100E-2020-065), and SUTD SKI Project (SKI 2021\_02\_06).
This work is also supported by TAILOR, a project funded by EU Horizon 2020 research and innovation programme under GA No 952215.

\normalsize

{\small
\bibliographystyleMain{ieee_fullname}
\bibliographyMain{egbib}
}

\newpage

\setlength{\textheight}{8.875in}
\setlength{\textwidth}{6.875in}
\setlength{\columnsep}{0.3125in}
\setlength{\topmargin}{0in}
\setlength{\headheight}{0in}
\setlength{\headsep}{0in}
\setlength{\parindent}{1pc}
\setlength{\oddsidemargin}{-.304in}
\setlength{\evensidemargin}{-.304in}

\title{\vspace{-0.2em} \Large \textbf{DiffPose: Toward More Reliable 3D Pose Estimation (Supplementary)} \vspace{0.3em} }

\author{Jia Gong\textsuperscript{1\dag}
~~~ Lin Geng Foo\textsuperscript{1\dag}
~~~ Zhipeng Fan\textsuperscript{2\S}
~~~ Qiuhong Ke\textsuperscript{3}
~~~ Hossein Rahmani\textsuperscript{4} ~~~ Jun Liu\textsuperscript{1\ddag} \\
\textsuperscript{1}Singapore University of Technology and Design ~~ \\
\textsuperscript{2}New York University ~~ 
\textsuperscript{3}Monash University ~~ 
\textsuperscript{4}Lancaster University\\
{\tt\small \{jia\_gong,lingeng\_foo\}@mymail.sutd.edu.sg, zf606@nyu.edu, qiuhong.ke@monash.edu, }\\ 
{\tt\small h.rahmani@lancaster.ac.uk, jun\_liu@sutd.edu.sg } \\
}

\maketitle

\setcounter{section}{0}
\setcounter{table}{0}
\setcounter{figure}{0}
\setcounter{equation}{0}

\section{Additional Details of GMM Forward Diffusion}

In Section 4.2 of the main paper, we describe the GMM-based forward diffusion process. Here, we explain it in more detail, particularly about how it can be framed in a step-wise formulation.
We first re-state Eq. 7 in the main paper as follows:
\begin{equation}
\begin{aligned}
\label{eq: modified diffusion linear function supp} 
\hat{h}_k = \mu^G + \sqrt{{\alpha}_k}(h_0- \mu^G) + \sqrt{(1-{\alpha_k})} \cdot \epsilon^G.
\end{aligned}
\end{equation}
where $\mu^G = \sum^M_{m=1} \textbf{1}_m \mu _m$, $\epsilon^G \sim \mathcal{N}(0, \sum^M_{m=1}(\textbf{1}_m \Sigma _m))$, and $\textbf{1}_m \in \{0,1\}$ is a binary indicator for the $m^{th}$ component such that $\sum^M_{m=1} \textbf{1}_m = 1$, and $Prob(\textbf{1}_m = 1) = \pi_m$.

We remark that Eq.~\ref{eq: modified diffusion linear function supp} directly formulates $\hat{h}_k$ as a function of $h_0$ instead of $\hat{h}_{k-1}$, 
because this clearly expresses the aim of our GMM-based forward diffusion design, i.e., such that the generated $\hat{h}_1,...,\hat{h}_K$ can converge to the fitted GMM model $\phi_{GMM}$.
Yet, we note that the step-wise formulation of $\hat{h}_k$ in terms of $\hat{h}_{k-1}$ can still be defined, if necessary. 
First, we sample according to probabilities $\{\pi_m  \}_{m=1}^M$, and select a Gaussian component $\hat{m}$, i.e., $\textbf{1}_{\hat{m}} =1$.
Next, we first calculate $\tilde{h}_0$, a ``centered" version of $h_0$, using $\tilde{h}_0 = h_0 - \mu^G$, where $\mu^G = \sum^M_{m=1}(\textbf{1}_m \mu_{m}) = \mu_{\hat{m}}$.
Then, we follow the step-wise formulation as follows:
\begin{equation}
\label{eq: standard diffusion step}
	 \tilde{h}_{k} = \sqrt{\frac{\alpha _k}{\alpha _{k-1}}} \tilde{h}_{k-1} + \sqrt{(1-\frac{\alpha _k}{\alpha _{k-1}})} \epsilon^G,
\end{equation}
where $\epsilon^G \sim \mathcal{N}(0, \sum^M_{m=1}(\textbf{1}_m \Sigma_m))$, which is equivalent to $\epsilon^G \sim \mathcal{N}(0, \Sigma_{\hat{m}})$.
After taking $k$ steps of Eq. \ref{eq: standard diffusion step} starting from $\tilde{h}_0$, we can get:

\begin{equation}
\setlength{\abovedisplayskip}{4pt}
\setlength{\belowdisplayskip}{4pt}
\begin{aligned}
\label{eq: standard diffusion step from h0} 
\tilde{h}_k = \sqrt{{\alpha}_k}(\tilde{h_0}) + \sqrt{(1-{\alpha_k})} \cdot \epsilon^G.
\end{aligned}
\end{equation}
We observe that the result of the stepwise formulation is thus equivalent to Eq.~\ref{eq: modified diffusion linear function supp}, as we can simply ``de-center" our $\tilde{h}_0$ and $ \tilde{h}_k$ by substituting $\tilde{h}_0 = h_0 - \mu^G$ and $\tilde{h}_k = \hat{h}_k - \mu^G$.

\section{Additional Details of Diffusion Network $g$} 
In order to provide information to the model regarding the current step number $k$, we generate a diffusion step embedding $f_D^k \in \mathbb{R}^{J \times 256}$ using the sinusoidal function. 
Specifically, at each even ($2j$) index of $f_D^k$, we set the element $f_D^k [2j]$ to $sin(k/10000^{2j/256})$, while at each odd ($2j+1$) index, we set the element $f_D^k [2j+1]$ to $cos(k/10000^{2j/256})$.

\begin{table*}[t]
\normalsize
\centering
\tabcolsep=5mm
\caption
{
Video-based results on Human3.6M with detected 2D poses in millimeters under P-MPJPE.
} 
\label{tab: Human3.6M video p2}
\vspace{-3mm}
\resizebox{0.96\textwidth}{!}{
\begin{tabular}{@{}l|ccccccccccccccc|c@{}}
\hline
\noalign{\smallskip}
P-MPJPE & Dir. & Disc. & Eat & Greet & Phone & Photo & Pose & Pur. & Sit & SitD. & Smoke & Wait & WalkD. & Walk & WalkT. & Avg \\
\noalign{\smallskip}
\hline
\noalign{\smallskip}
Lin \cite{lin2019trajectory2} &
32.5&35.3&34.3&36.2&37.8&43.0&33.0&32.2&45.7&51.8&38.4&32.8&37.5&25.8&28.9&36.8 \\
Pavllo \cite{pavllo20193d2} & 34.1&36.1&34.4&37.2&36.4&42.2&34.4&33.6&45.0&52.5&37.4&33.8&37.8&25.6&27.3&36.5 \\
Liu \cite{liu2020attention2}& 32.3&{35.2}&33.3&35.8&35.9&41.5&33.2&32.7&44.6&50.9&37.0&{32.4}&37.0&25.2&27.2&35.6 \\
Zheng \textit{et al.}~\cite{zheng20213d2} &
32.5&{34.8}&{32.6}&34.6&{35.3}&{39.5}&32.1&32.0&42.8&48.5&{34.8}&{32.4}&35.3&24.5&26.0&{34.6}\\
Li \cite{li2022mhformer2} & 31.5 & 34.9 & 32.8 & 33.6 & 35.3 & 39.6 & 32.0 & 32.2 & 43.5 & 48.7 & 36.4 & 32.6 & 34.3 & 23.9 & 25.1 & 34.4 \\
Zhang \cite{zhang2022mixste2} & \underline{28.0} & \underline{30.9} & \underline{28.6} & \underline{30.7} & \underline{30.4} & \textbf{34.6} & \underline{28.6} & \underline{28.1} & \underline{37.1} & \underline{47.3} & \underline{30.5} & \underline{29.7} & \underline{30.5} & \underline{21.6} & \underline{20.0} & \underline{30.6} \\
\hline
ours & \textbf{26.3} & \textbf{29.0} & \textbf{26.1} & \textbf{27.8} & \textbf{28.4} & \textbf{34.6} & \textbf{26.9} & \textbf{26.5} & \textbf{36.8} & \textbf{39.2} & \textbf{29.4} & \textbf{26.8} & \textbf{28.4} & \textbf{18.6} & \textbf{19.2} & \textbf{28.7} \\
\hline
\noalign{\smallskip}
\end{tabular}
}
\end{table*}

\begin{table*}[t]
\normalsize
\centering
\tabcolsep=5mm
\caption
{Frame-based results on Human3.6M with detected 2D poses in millimeters under P-MPJPE.
} 
\label{tab: Human3.6M frame p2}
\vspace{-3mm}
\resizebox{0.96\textwidth}{!}{
\begin{tabular}{@{}l|ccccccccccccccc|c@{}}
\hline
\noalign{\smallskip}
P-MPJPE & Dir. & Disc. & Eat & Greet & Phone & Photo & Pose & Pur. & Sit & SitD. & Smoke & Wait & WalkD. & Walk & WalkT. & Avg \\
\noalign{\smallskip}
\hline
\noalign{\smallskip}
Sun \cite{sun2017compositional2} & 42.1 & 44.3 & 45.0 & 45.4 & 51.5 & 53.0 & 43.2 & 41.3 & 59.3 & 73.3 & 51.0 & 44.0 & 48.0 & 38.3 & 44.8 & 48.3 \\
Martinez \cite{martinez2017simple2} & 39.5 & 43.2 & 46.4 & 47.0 & 51.0 & 56.0 & 41.4 & 40.6 & 56.5 & 69.4 & 49.2 & 45.0 & 49.5 & 38.0 & 43.1 & 47.7 \\
Pavlakos \cite{pavlakos2017coarse2} & \underline{34.7} & \underline{39.8} & 41.8 & \textbf{38.6} & \underline{42.5} & \underline{47.5} & 38.0 & 36.6 & 50.7 & 56.8 & 42.6 & 39.6 & 43.9 & 32.1 & 36.5 & 41.8 \\
Liu \cite{liu2020comprehensive2} & {35.9} & {40.0} & \underline{38.0} & {41.5} & \underline{42.5} & {51.4} & \underline{37.8} & \underline{36.0} & \underline{48.6} & \underline{56.6} & \underline{41.8} & \underline{38.3} & \underline{42.7} & \underline{31.7} & \underline{36.2} & \underline{41.2} \\
\hline
ours & \textbf{33.9} & \textbf{38.2} & \textbf{36.0}  & \underline{39.2}  & \textbf{40.2}  & \textbf{46.5}  & \textbf{35.8}  &  \textbf{34.8} &  \textbf{48.0} & \textbf{52.5}  & \textbf{41.2}  &  \textbf{36.5} &  \textbf{40.9} &  \textbf{30.3} &  \textbf{33.8} & \textbf{39.2} \\
\hline
\noalign{\smallskip}
\end{tabular}
}
\end{table*}

\section{More Implementation Details}
In the forward diffusion process, we generate the decreasing sequence $\alpha _{1:K}$ via the formula:
$\alpha_k = \prod ^{k}_{i=1}(1-{\beta_i}),$
where $\beta _{1:K}$ is a sequence from $1e-4$ to $2e-3$, which is interpolated by the linear function.
To optimize the GMM parameters $\phi_{GMM}$, we sample 1000 poses from $H_K$ (i.e., $N_{GMM}=1000$) and then model $H_K$ via a GMM model.

During model pre-training, the Context Encoder $\phi_{ST}$ is first pre-trained on the training set to predict 3D poses from 2D poses.
Then we adopt the Adam optimizer \cite{kingma2014adam} to train our diffusion model $g$, where the initial learning rate is set to $1e{-4}$ with a decay rate of $0.9$ after ten epochs, and the batch size is set to $4096$. 
Our DiffPose is implemented using PyTorch, and can be trained on a single GeForce RTX 3090 GPU within 96 hours. 

\begin{figure}[t]
  \centering
  \includegraphics[width=1.00\linewidth]{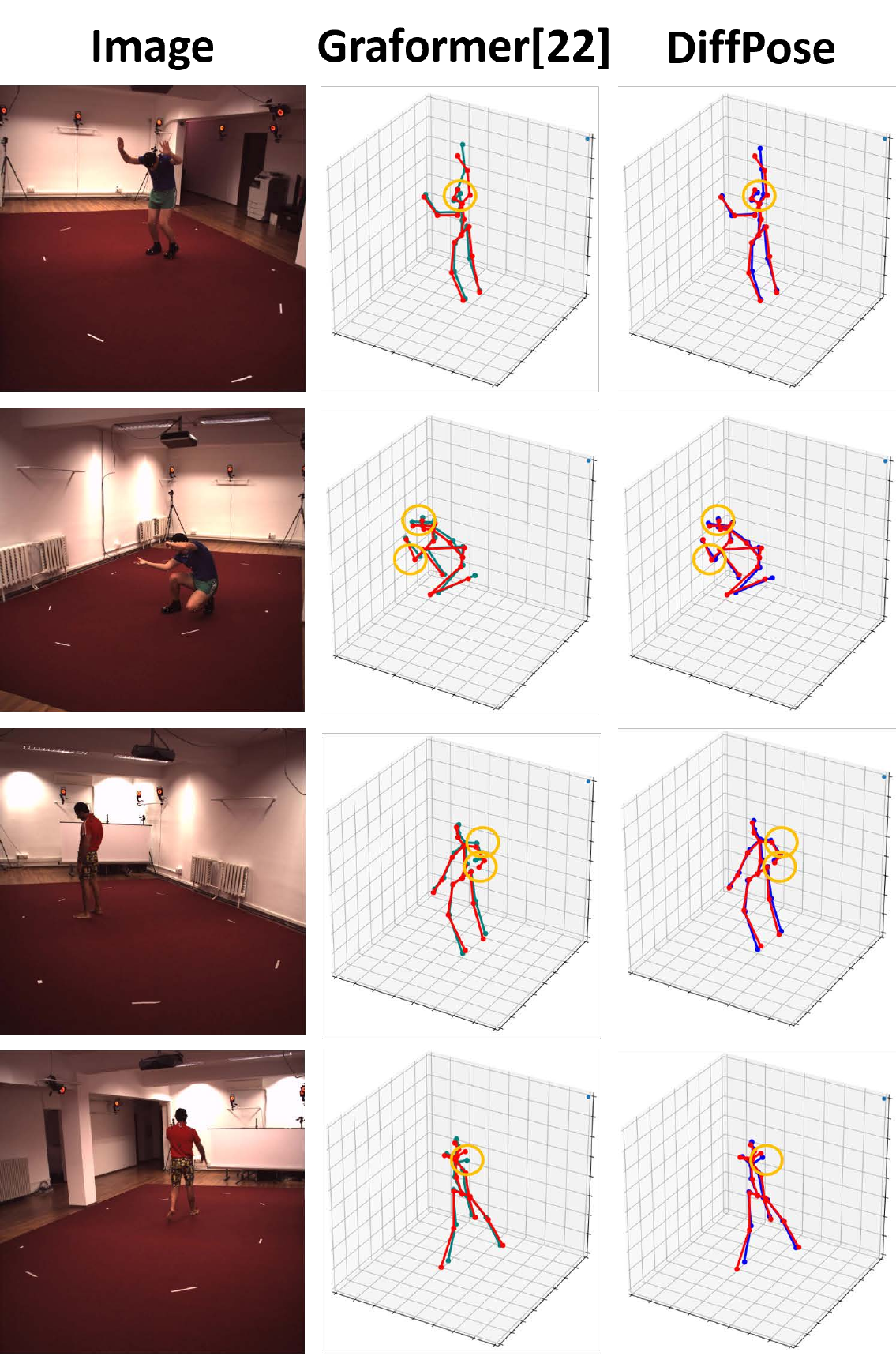}
  \vspace{-7mm}
  \caption{Qualitative comparison between Graformer \cite{zhao2022graformer2} and our method. Red colored 3D pose corresponds to the ground truth. 
  }
  \label{fig: additional qualitative result}
\end{figure}

\section{Experiment Results on Human3.6M under P-MPJPE (Protocol 2)}
Tab.~\ref{tab: Human3.6M video p2} and Tab.~\ref{tab: Human3.6M frame p2} present the video-based and frame-based results of our DiffPose on Human3.6M under P-MPJPE, where the input 2D poses are detected by CPN \cite{chen2018cascaded2}. 
As shown in Tab.~\ref{tab: Human3.6M video p2}, our DiffPose can significantly outperform the state-of-the-art methods \cite{li2022mhformer2,zhang2022mixste2} on all actions with a large margin. Moreover, from Tab.~\ref{tab: Human3.6M frame p2}, we observe that our method can achieve promising performance on the challenging frame-based setting.

\section{Additional Results}
In this section, we further investigate the performance of our method on the frame-based scenario, by conducting experiments on Human3.6M \cite{ionescu2013human32}. 

\textbf{3D Pose visualization.} First, we qualitatively compare our method with state-of-the-art method \cite{zhao2022graformer2} in this setting, and present results in Fig.~\ref{fig: additional qualitative result}. 
We observe that our method can predict more reliable and accurate poses, especially for novel human gestures (e.g., the first and second rows in Fig.~\ref{fig: additional qualitative result}) and occluded body parts (e.g., the third and fourth rows in  Fig.~\ref{fig: additional qualitative result}).

\begin{figure}[t]
  \centering
  \includegraphics[width=1.00\linewidth]{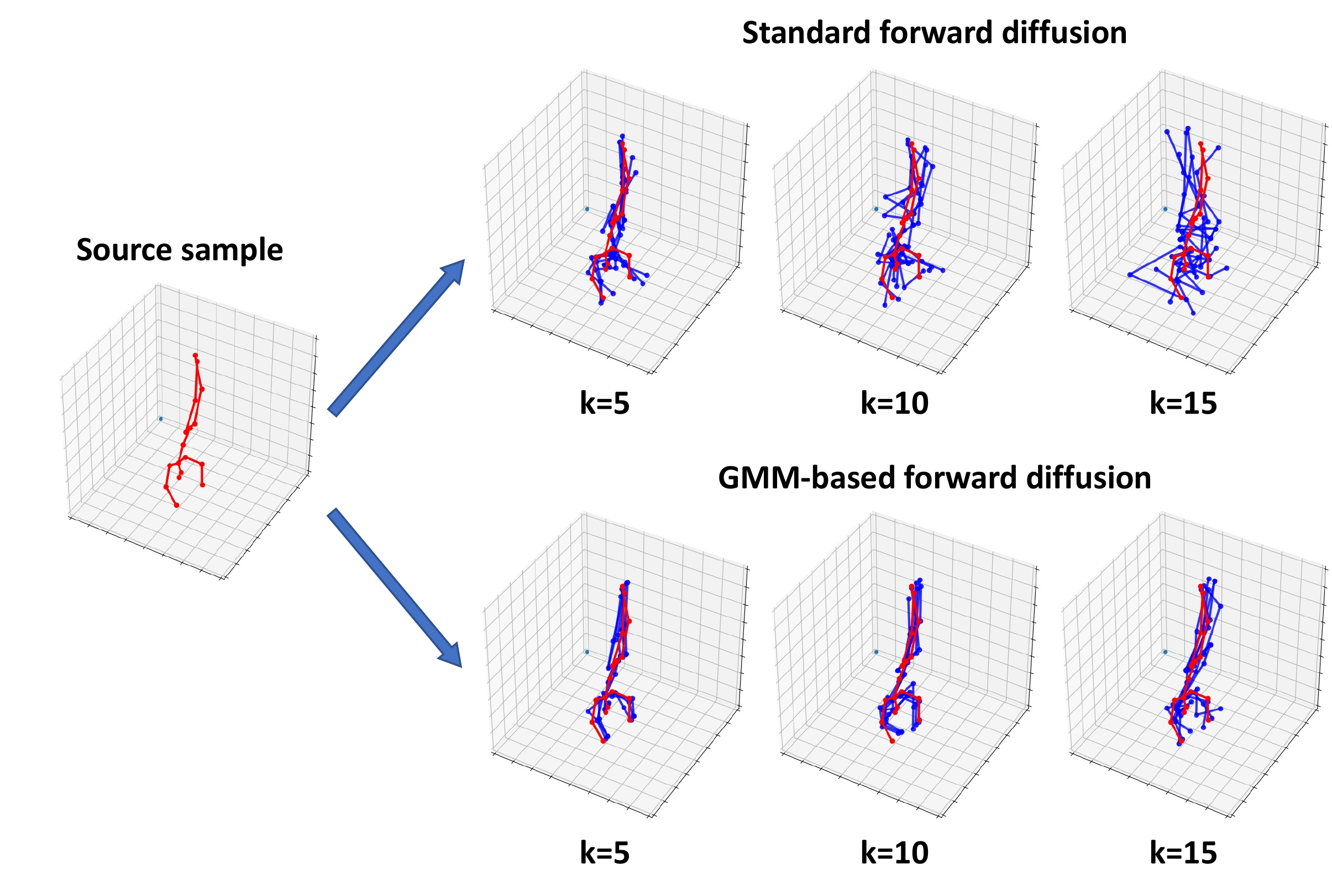}
  \vspace{-7mm}
  \caption{Qualitative comparison between standard diffusion forward process and our GMM-based forward diffusion process.}
  \label{fig: supp forward}
\end{figure}

\textbf{Forward diffusion process visualization.} 
Extending from our results in Tab.~5 of the main paper, here we qualitatively compare our GMM-based forward diffusion process with the standard diffusion process (as described in Sec.~3 of our main paper). 
As shown in Fig.~\ref{fig: supp forward}, the standard diffusion process recurrently adds noise to the source sample and tends to spread the joints' positions to the whole space. 
However, our GMM-based diffusion process can add noise according to pose-specific information (obtained from heatmaps) and the data distribution, which generates noise in a more constrained manner.
Thus, during training, the GMM-based diffusion process allows us to initialize a $\hat{H}_K$ that captures the uncertainty of the 3D pose, which boosts the performance of DiffPose.

\begin{figure}[t]
  \centering
  \includegraphics[width=1.00\linewidth]{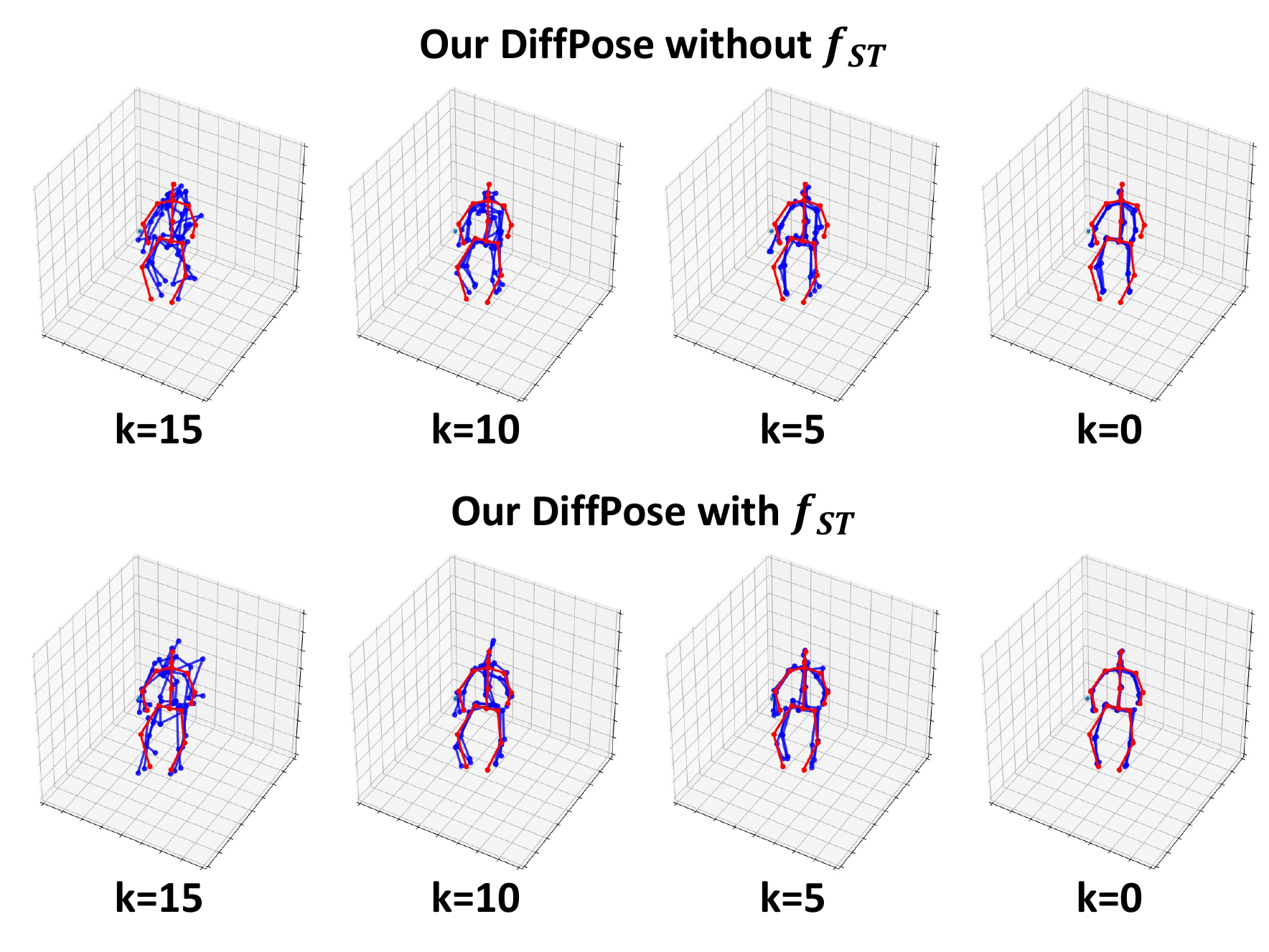}
  \vspace{-7mm}
  \caption{
  Qualitative comparison between our reverse diffusion process conditioned on context information $f_{ST}$ (bottom), against a reverse diffusion process without using $f_{ST}$ (top).
  }
  \label{fig: supp reverse}
\end{figure}

\textbf{Reverse diffusion process visualization.} 
We visualize the poses reconstructed by our diffusion model with/without the context information $f_{ST}$. Note that the model without $f_{ST}$ means that no context decoder is used. From the last column of Fig.~\ref{fig: supp reverse}, we observe that both methods can reconstruct realistic human poses while the model with $f_{ST}$ can predict more accurate poses. Moreover, compared to the unconditioned reverse diffusion process (i.e., the model without $f_{ST}$), the model conditioned by $f_{ST}$ can converge to the desired pose faster.

\section{Future Work}
In this work, we explore a novel diffusion-based framework to tackle monocular 3D pose estimation.
Future work includes more investigations into the architecture of the diffusion network, as well as extending to the online setting \cite{foo2023system2,wu2019liteeval,habibian2021skip}, the few-shot setting \cite{Zhao_2023_tip_fsc,taylor2012vitruvian} and other pose-based tasks \cite{foo2023unified2,foo2022era,shi2019two,yan2018spatial,liu2016spatio}.

{\small
\bibliographystyle{ieee_fullname}
\bibliography{egbib_supp}

\begin{thebibliography}{10}\itemsep=-1pt

\bibitem{austin2021structured}
Jacob Austin, Daniel~D Johnson, Jonathan Ho, Daniel Tarlow, and Rianne van~den
  Berg.
\newblock Structured denoising diffusion models in discrete state-spaces.
\newblock {\em Advances in Neural Information Processing Systems},
  34:17981--17993, 2021.

\bibitem{cai2019exploiting}
Yujun Cai, Liuhao Ge, Jun Liu, Jianfei Cai, Tat-Jen Cham, Junsong Yuan, and
  Nadia~Magnenat Thalmann.
\newblock Exploiting spatial-temporal relationships for 3d pose estimation via
  graph convolutional networks.
\newblock In {\em Proceedings of the IEEE/CVF international conference on
  computer vision}, pages 2272--2281, 2019.

\bibitem{chen2021anatomy}
Tianlang Chen, Chen Fang, Xiaohui Shen, Yiheng Zhu, Zhili Chen, and Jiebo Luo.
\newblock Anatomy-aware 3d human pose estimation with bone-based pose
  decomposition.
\newblock {\em IEEE Transactions on Circuits and Systems for Video Technology},
  32(1):198--209, 2021.

\bibitem{chen2018cascaded}
Yilun Chen, Zhicheng Wang, Yuxiang Peng, Zhiqiang Zhang, Gang Yu, and Jian Sun.
\newblock Cascaded pyramid network for multi-person pose estimation.
\newblock In {\em Proceedings of the IEEE conference on computer vision and
  pattern recognition}, pages 7103--7112, 2018.

\bibitem{chessa2019grasping}
Manuela Chessa, Guido Maiello, Lina~K Klein, Vivian~C Paulun, and Fabio Solari.
\newblock Grasping objects in immersive virtual reality.
\newblock In {\em 2019 IEEE Conference on Virtual Reality and 3D User
  Interfaces (VR)}, pages 1749--1754. IEEE, 2019.

\bibitem{ci2019optimizing}
Hai Ci, Chunyu Wang, Xiaoxuan Ma, and Yizhou Wang.
\newblock Optimizing network structure for 3d human pose estimation.
\newblock In {\em Proceedings of the IEEE/CVF international conference on
  computer vision}, pages 2262--2271, 2019.

\bibitem{fan2020adaptive}
Zhipeng Fan, Jun Liu, and Yao Wang.
\newblock Adaptive computationally efficient network for monocular 3d hand pose
  estimation.
\newblock In {\em Computer Vision--ECCV 2020: 16th European Conference,
  Glasgow, UK, August 23--28, 2020, Proceedings, Part IV 16}, pages 127--144.
  Springer, 2020.

\bibitem{fan2021motion}
Zhipeng Fan, Jun Liu, and Yao Wang.
\newblock Motion adaptive pose estimation from compressed videos.
\newblock In {\em Proceedings of the IEEE/CVF International Conference on
  Computer Vision}, pages 11719--11728, 2021.

\bibitem{foo2023system}
Lin~Geng Foo, Jia Gong, Zhipeng Fan, and Jun Liu.
\newblock System-status-aware adaptive network for online streaming video
  understanding.
\newblock In {\em Proceedings of the IEEE/CVF Conference on Computer Vision and
  Pattern Recognition (CVPR)}, June 2023.

\bibitem{foo2023unified}
Lin~Geng Foo, Tianjiao Li, Hossein Rahmani, Qiuhong Ke, and Jun Liu.
\newblock Unified pose sequence modeling.
\newblock In {\em Proceedings of the IEEE/CVF Conference on Computer Vision and
  Pattern Recognition (CVPR)}, June 2023.

\bibitem{gu2022stochastic}
Tianpei Gu, Guangyi Chen, Junlong Li, Chunze Lin, Yongming Rao, Jie Zhou, and
  Jiwen Lu.
\newblock Stochastic trajectory prediction via motion indeterminacy diffusion.
\newblock In {\em Proceedings of the IEEE/CVF Conference on Computer Vision and
  Pattern Recognition}, pages 17113--17122, 2022.

\bibitem{ho2020denoising}
Jonathan Ho, Ajay Jain, and Pieter Abbeel.
\newblock Denoising diffusion probabilistic models.
\newblock {\em Advances in Neural Information Processing Systems},
  33:6840--6851, 2020.

\bibitem{hossain2018exploiting}
Mir Rayat~Imtiaz Hossain and James~J Little.
\newblock Exploiting temporal information for 3d human pose estimation.
\newblock In {\em ECCV}, pages 68--84, 2018.

\bibitem{hu2021conditional}
Wenbo Hu, Changgong Zhang, Fangneng Zhan, Lei Zhang, and Tien-Tsin Wong.
\newblock Conditional directed graph convolution for 3d human pose estimation.
\newblock In {\em Proceedings of the 29th ACM International Conference on
  Multimedia}, pages 602--611, 2021.

\bibitem{ionescu2013human3}
Catalin Ionescu, Dragos Papava, Vlad Olaru, and Cristian Sminchisescu.
\newblock Human3.6m: Large scale datasets and predictive methods for 3d human
  sensing in natural environments.
\newblock {\em IEEE transactions on pattern analysis and machine intelligence},
  36(7):1325--1339, 2013.

\bibitem{khirodkar2021multi}
Rawal Khirodkar, Visesh Chari, Amit Agrawal, and Ambrish Tyagi.
\newblock Multi-instance pose networks: Rethinking top-down pose estimation.
\newblock In {\em Proceedings of the IEEE/CVF International Conference on
  Computer Vision}, pages 3122--3131, 2021.

\bibitem{li2019generating}
Chen Li and Gim~Hee Lee.
\newblock Generating multiple hypotheses for 3d human pose estimation with
  mixture density network.
\newblock In {\em Proceedings of the IEEE/CVF conference on computer vision and
  pattern recognition}, pages 9887--9895, 2019.

\bibitem{li1999mixture}
Jonathan Li and Andrew Barron.
\newblock Mixture density estimation.
\newblock {\em Advances in neural information processing systems}, 12, 1999.

\bibitem{li2022mhformer}
Wenhao Li, Hong Liu, Hao Tang, Pichao Wang, and Luc Van~Gool.
\newblock Mhformer: Multi-hypothesis transformer for 3d human pose estimation.
\newblock In {\em Proceedings of the IEEE/CVF Conference on Computer Vision and
  Pattern Recognition}, pages 13147--13156, 2022.

\bibitem{li2022diffusion}
Xiang~Lisa Li, John Thickstun, Ishaan Gulrajani, Percy Liang, and Tatsunori~B
  Hashimoto.
\newblock Diffusion-lm improves controllable text generation.
\newblock {\em arXiv preprint arXiv:2205.14217}, 2022.

\bibitem{liang2020multi}
Xing Liang, Anastassia Angelopoulou, Epaminondas Kapetanios, Bencie Woll, Reda
  Al~Batat, and Tyron Woolfe.
\newblock A multi-modal machine learning approach and toolkit to automate
  recognition of early stages of dementia among british sign language users.
\newblock In {\em European Conference on Computer Vision}, pages 278--293.
  Springer, 2020.

\bibitem{lin2019trajectory}
Jiahao Lin and Gim~Hee Lee.
\newblock Trajectory space factorization for deep video-based 3d human pose
  estimation.
\newblock In {\em BMVC}, 2019.

\bibitem{liu2020comprehensive}
Kenkun Liu, Rongqi Ding, Zhiming Zou, Le Wang, and Wei Tang.
\newblock A comprehensive study of weight sharing in graph networks for 3d
  human pose estimation.
\newblock In {\em European Conference on Computer Vision}, pages 318--334.
  Springer, 2020.

\bibitem{liu2020attention}
Ruixu Liu, Ju Shen, He Wang, Chen Chen, Sen-ching Cheung, and Vijayan Asari.
\newblock Attention mechanism exploits temporal contexts: Real-time 3d human
  pose reconstruction.
\newblock In {\em Proceedings of the IEEE/CVF Conference on Computer Vision and
  Pattern Recognition}, pages 5064--5073, 2020.

\bibitem{lugmayr2022repaint}
Andreas Lugmayr, Martin Danelljan, Andres Romero, Fisher Yu, Radu Timofte, and
  Luc Van~Gool.
\newblock Repaint: Inpainting using denoising diffusion probabilistic models.
\newblock In {\em Proceedings of the IEEE/CVF Conference on Computer Vision and
  Pattern Recognition}, pages 11461--11471, 2022.

\bibitem{martinez2017simple}
Julieta Martinez, Rayat Hossain, Javier Romero, and James~J Little.
\newblock A simple yet effective baseline for 3d human pose estimation.
\newblock In {\em IEEE ICCV}, pages 2640--2649, 2017.

\bibitem{mehta2017monocular}
Dushyant Mehta, Helge Rhodin, Dan Casas, Pascal Fua, Oleksandr Sotnychenko,
  Weipeng Xu, and Christian Theobalt.
\newblock Monocular 3d human pose estimation in the wild using improved cnn
  supervision.
\newblock In {\em 2017 international conference on 3D vision (3DV)}, pages
  506--516. IEEE, 2017.

\bibitem{nachmani2021non}
Eliya Nachmani, Robin~San Roman, and Lior Wolf.
\newblock Non gaussian denoising diffusion models.
\newblock {\em arXiv preprint arXiv:2106.07582}, 2021.

\bibitem{nichol2021improved}
Alexander~Quinn Nichol and Prafulla Dhariwal.
\newblock Improved denoising diffusion probabilistic models.
\newblock In {\em International Conference on Machine Learning}, pages
  8162--8171. PMLR, 2021.

\bibitem{park20163d}
Sungheon Park, Jihye Hwang, and Nojun Kwak.
\newblock 3d human pose estimation using convolutional neural networks with 2d
  pose information.
\newblock In {\em European Conference on Computer Vision}, pages 156--169.
  Springer, 2016.

\bibitem{pavlakos2017coarse}
Georgios Pavlakos, Xiaowei Zhou, Konstantinos~G Derpanis, and Kostas
  Daniilidis.
\newblock Coarse-to-fine volumetric prediction for single-image 3d human pose.
\newblock In {\em IEEE CVPR}, pages 7025--7034, 2017.

\bibitem{pavllo20193d}
Dario Pavllo, Christoph Feichtenhofer, David Grangier, and Michael Auli.
\newblock 3d human pose estimation in video with temporal convolutions and
  semi-supervised training.
\newblock In {\em Proceedings of the IEEE/CVF Conference on Computer Vision and
  Pattern Recognition}, pages 7753--7762, 2019.

\bibitem{rombach2022high}
Robin Rombach, Andreas Blattmann, Dominik Lorenz, Patrick Esser, and Bj{\"o}rn
  Ommer.
\newblock High-resolution image synthesis with latent diffusion models.
\newblock In {\em Proceedings of the IEEE/CVF Conference on Computer Vision and
  Pattern Recognition}, pages 10684--10695, 2022.

\bibitem{shan2022p}
Wenkang Shan, Zhenhua Liu, Xinfeng Zhang, Shanshe Wang, Siwei Ma, and Wen Gao.
\newblock P-stmo: Pre-trained spatial temporal many-to-one model for 3d human
  pose estimation.
\newblock In {\em ECCV}, page 461–478, 2022.

\bibitem{shan2021improving}
Wenkang Shan, Haopeng Lu, Shanshe Wang, Xinfeng Zhang, and Wen Gao.
\newblock Improving robustness and accuracy via relative information encoding
  in 3d human pose estimation.
\newblock In {\em Proceedings of the 29th ACM International Conference on
  Multimedia}, pages 3446--3454, 2021.

\bibitem{sharma2019monocular}
Saurabh Sharma, Pavan~Teja Varigonda, Prashast Bindal, Abhishek Sharma, and
  Arjun Jain.
\newblock Monocular 3d human pose estimation by generation and ordinal ranking.
\newblock In {\em Proceedings of the IEEE/CVF international conference on
  computer vision}, pages 2325--2334, 2019.

\bibitem{sohl2015deep}
Jascha Sohl-Dickstein, Eric Weiss, Niru Maheswaranathan, and Surya Ganguli.
\newblock Deep unsupervised learning using nonequilibrium thermodynamics.
\newblock In {\em International Conference on Machine Learning}, pages
  2256--2265. PMLR, 2015.

\bibitem{song2021denoising}
Jiaming Song, Chenlin Meng, and Stefano Ermon.
\newblock Denoising diffusion implicit models.
\newblock In {\em International Conference on Learning Representations}, 2021.

\bibitem{song2019generative}
Yang Song and Stefano Ermon.
\newblock Generative modeling by estimating gradients of the data distribution.
\newblock {\em Advances in Neural Information Processing Systems}, 32, 2019.

\bibitem{sridhar2015investigating}
Srinath Sridhar, Anna~Maria Feit, Christian Theobalt, and Antti Oulasvirta.
\newblock Investigating the dexterity of multi-finger input for mid-air text
  entry.
\newblock In {\em Proceedings of the 33rd Annual ACM Conference on Human
  Factors in Computing Systems}, pages 3643--3652, 2015.

\bibitem{sun2017compositional}
Xiao Sun, Jiaxiang Shang, Shuang Liang, and Yichen Wei.
\newblock Compositional human pose regression.
\newblock In {\em IEEE ICCV}, pages 2602--2611, 2017.

\bibitem{sun2018integral}
Xiao Sun, Bin Xiao, Fangyin Wei, Shuang Liang, and Yichen Wei.
\newblock Integral human pose regression.
\newblock In {\em Proceedings of the European conference on computer vision
  (ECCV)}, pages 529--545, 2018.

\bibitem{wang2022low}
Chen Wang, Feng Zhang, Xiatian Zhu, and Shuzhi~Sam Ge.
\newblock Low-resolution human pose estimation.
\newblock {\em Pattern Recognition}, 126:108579, 2022.

\bibitem{wang2020motion}
Jingbo Wang, Sijie Yan, Yuanjun Xiong, and Dahua Lin.
\newblock Motion guided 3d pose estimation from videos.
\newblock In {\em European Conference on Computer Vision}, pages 764--780.
  Springer, 2020.

\bibitem{xu2020deep}
Jingwei Xu, Zhenbo Yu, Bingbing Ni, Jiancheng Yang, Xiaokang Yang, and Wenjun
  Zhang.
\newblock Deep kinematics analysis for monocular 3d human pose estimation.
\newblock In {\em Proceedings of the IEEE/CVF Conference on Computer Vision and
  Pattern Recognition}, pages 899--908, 2020.

\bibitem{xu2021graph}
Tianhan Xu and Wataru Takano.
\newblock Graph stacked hourglass networks for 3d human pose estimation.
\newblock In {\em Proceedings of the IEEE/CVF conference on computer vision and
  pattern recognition}, pages 16105--16114, 2021.

\bibitem{yang20183d}
Wei Yang, Wanli Ouyang, Xiaolong Wang, Jimmy Ren, Hongsheng Li, and Xiaogang
  Wang.
\newblock 3d human pose estimation in the wild by adversarial learning.
\newblock In {\em IEEE CVPR}, pages 5255--5264, 2018.

\bibitem{zeng2020srnet}
Ailing Zeng, Xiao Sun, Fuyang Huang, Minhao Liu, Qiang Xu, and Stephen Lin.
\newblock Srnet: Improving generalization in 3d human pose estimation with a
  split-and-recombine approach.
\newblock In {\em European Conference on Computer Vision}, pages 507--523.
  Springer, 2020.

\bibitem{zhang2022mixste}
Jinlu Zhang, Zhigang Tu, Jianyu Yang, Yujin Chen, and Junsong Yuan.
\newblock Mixste: Seq2seq mixed spatio-temporal encoder for 3d human pose
  estimation in video.
\newblock In {\em Proceedings of the IEEE/CVF Conference on Computer Vision and
  Pattern Recognition}, pages 13232--13242, 2022.

\bibitem{zhaoCVPR19semantic}
Long Zhao, Xi Peng, Yu Tian, Mubbasir Kapadia, and Dimitris~N. Metaxas.
\newblock Semantic graph convolutional networks for 3d human pose regression.
\newblock In {\em IEEE Conference on Computer Vision and Pattern Recognition
  (CVPR)}, pages 3425--3435, 2019.

\bibitem{zhao2019semantic}
Long Zhao, Xi Peng, Yu Tian, Mubbasir Kapadia, and Dimitris~N Metaxas.
\newblock Semantic graph convolutional networks for 3d human pose regression.
\newblock In {\em IEEE CVPR}, pages 3425--3435, 2019.

\bibitem{zhao2022graformer}
Weixi Zhao, Weiqiang Wang, and Yunjie Tian.
\newblock Graformer: Graph-oriented transformer for 3d pose estimation.
\newblock In {\em Proceedings of the IEEE/CVF Conference on Computer Vision and
  Pattern Recognition}, pages 20438--20447, 2022.

\bibitem{Zhao_2023_arxiv_watermark_dm}
Yunqing Zhao, Tianyu Pang, Chao Du, Xiao Yang, Ngai-Man Cheung, and Min Lin.
\newblock A recipe for watermarking diffusion models.
\newblock {\em arXiv preprint arXiv: 2303.10137}, 2023.

\bibitem{zheng20213d}
Ce Zheng, Sijie Zhu, Matias Mendieta, Taojiannan Yang, Chen Chen, and Zhengming
  Ding.
\newblock 3d human pose estimation with spatial and temporal transformers.
\newblock In {\em Proceedings of the IEEE/CVF International Conference on
  Computer Vision}, pages 11656--11665, 2021.

\end{thebibliography}


\begin{thebibliography}{10}\itemsep=-1pt

\bibitem{chen2018cascaded2}
Yilun Chen, Zhicheng Wang, Yuxiang Peng, Zhiqiang Zhang, Gang Yu, and Jian Sun.
\newblock Cascaded pyramid network for multi-person pose estimation.
\newblock In {\em Proceedings of the IEEE conference on computer vision and
  pattern recognition}, pages 7103--7112, 2018.

\bibitem{foo2023system2}
Lin~Geng Foo, Jia Gong, Zhipeng Fan, and Jun Liu.
\newblock System-status-aware adaptive network for online streaming video
  understanding.
\newblock In {\em Proceedings of the IEEE/CVF Conference on Computer Vision and
  Pattern Recognition (CVPR)}, June 2023.

\bibitem{foo2022era}
Lin~Geng Foo, Tianjiao Li, Hossein Rahmani, Qiuhong Ke, and Jun Liu.
\newblock Era: Expert retrieval and assembly for early action prediction.
\newblock In {\em Computer Vision--ECCV 2022: 17th European Conference, Tel
  Aviv, Israel, October 23--27, 2022, Proceedings, Part XXXIV}, pages 670--688.
  Springer, 2022.

\bibitem{foo2023unified2}
Lin~Geng Foo, Tianjiao Li, Hossein Rahmani, Qiuhong Ke, and Jun Liu.
\newblock Unified pose sequence modeling.
\newblock In {\em Proceedings of the IEEE/CVF Conference on Computer Vision and
  Pattern Recognition (CVPR)}, June 2023.

\bibitem{habibian2021skip}
Amirhossein Habibian, Davide Abati, Taco~S Cohen, and Babak~Ehteshami Bejnordi.
\newblock Skip-convolutions for efficient video processing.
\newblock In {\em Proceedings of the IEEE/CVF Conference on Computer Vision and
  Pattern Recognition}, pages 2695--2704, 2021.

\bibitem{ionescu2013human32}
Catalin Ionescu, Dragos Papava, Vlad Olaru, and Cristian Sminchisescu.
\newblock Human3.6m: Large scale datasets and predictive methods for 3d human
  sensing in natural environments.
\newblock {\em IEEE transactions on pattern analysis and machine intelligence},
  36(7):1325--1339, 2013.

\bibitem{kingma2014adam}
Diederik~P Kingma and Jimmy Ba.
\newblock Adam: A method for stochastic optimization.
\newblock {\em arXiv preprint arXiv:1412.6980}, 2014.

\bibitem{li2022mhformer2}
Wenhao Li, Hong Liu, Hao Tang, Pichao Wang, and Luc Van~Gool.
\newblock Mhformer: Multi-hypothesis transformer for 3d human pose estimation.
\newblock In {\em Proceedings of the IEEE/CVF Conference on Computer Vision and
  Pattern Recognition}, pages 13147--13156, 2022.

\bibitem{lin2019trajectory2}
Jiahao Lin and Gim~Hee Lee.
\newblock Trajectory space factorization for deep video-based 3d human pose
  estimation.
\newblock In {\em BMVC}, 2019.

\bibitem{liu2016spatio}
Jun Liu, Amir Shahroudy, Dong Xu, and Gang Wang.
\newblock Spatio-temporal lstm with trust gates for 3d human action
  recognition.
\newblock In {\em European conference on computer vision}, pages 816--833.
  Springer, 2016.

\bibitem{liu2020comprehensive2}
Kenkun Liu, Rongqi Ding, Zhiming Zou, Le Wang, and Wei Tang.
\newblock A comprehensive study of weight sharing in graph networks for 3d
  human pose estimation.
\newblock In {\em European Conference on Computer Vision}, pages 318--334.
  Springer, 2020.

\bibitem{liu2020attention2}
Ruixu Liu, Ju Shen, He Wang, Chen Chen, Sen-ching Cheung, and Vijayan Asari.
\newblock Attention mechanism exploits temporal contexts: Real-time 3d human
  pose reconstruction.
\newblock In {\em Proceedings of the IEEE/CVF Conference on Computer Vision and
  Pattern Recognition}, pages 5064--5073, 2020.

\bibitem{martinez2017simple2}
Julieta Martinez, Rayat Hossain, Javier Romero, and James~J Little.
\newblock A simple yet effective baseline for 3d human pose estimation.
\newblock In {\em IEEE ICCV}, pages 2640--2649, 2017.

\bibitem{pavlakos2017coarse2}
Georgios Pavlakos, Xiaowei Zhou, Konstantinos~G Derpanis, and Kostas
  Daniilidis.
\newblock Coarse-to-fine volumetric prediction for single-image 3d human pose.
\newblock In {\em IEEE CVPR}, pages 7025--7034, 2017.

\bibitem{pavllo20193d2}
Dario Pavllo, Christoph Feichtenhofer, David Grangier, and Michael Auli.
\newblock 3d human pose estimation in video with temporal convolutions and
  semi-supervised training.
\newblock In {\em Proceedings of the IEEE/CVF Conference on Computer Vision and
  Pattern Recognition}, pages 7753--7762, 2019.

\bibitem{shi2019two}
Lei Shi, Yifan Zhang, Jian Cheng, and Hanqing Lu.
\newblock Two-stream adaptive graph convolutional networks for skeleton-based
  action recognition.
\newblock In {\em Proceedings of the IEEE/CVF conference on computer vision and
  pattern recognition}, pages 12026--12035, 2019.

\bibitem{sun2017compositional2}
Xiao Sun, Jiaxiang Shang, Shuang Liang, and Yichen Wei.
\newblock Compositional human pose regression.
\newblock In {\em IEEE ICCV}, pages 2602--2611, 2017.

\bibitem{taylor2012vitruvian}
Jonathan Taylor, Jamie Shotton, Toby Sharp, and Andrew Fitzgibbon.
\newblock The vitruvian manifold: Inferring dense correspondences for one-shot
  human pose estimation.
\newblock In {\em 2012 IEEE Conference on Computer Vision and Pattern
  Recognition}, pages 103--110. IEEE, 2012.

\bibitem{wu2019liteeval}
Zuxuan Wu, Caiming Xiong, Yu-Gang Jiang, and Larry~S Davis.
\newblock Liteeval: A coarse-to-fine framework for resource efficient video
  recognition.
\newblock In H. Wallach, H. Larochelle, A. Beygelzimer, F. d\textquotesingle
  Alch\'{e}-Buc, E. Fox, and R. Garnett, editors, {\em Advances in Neural
  Information Processing Systems}, volume~32. Curran Associates, Inc., 2019.

\bibitem{yan2018spatial}
Sijie Yan, Yuanjun Xiong, and Dahua Lin.
\newblock Spatial temporal graph convolutional networks for skeleton-based
  action recognition.
\newblock In {\em Thirty-second AAAI conference on artificial intelligence},
  2018.

\bibitem{zhang2022mixste2}
Jinlu Zhang, Zhigang Tu, Jianyu Yang, Yujin Chen, and Junsong Yuan.
\newblock Mixste: Seq2seq mixed spatio-temporal encoder for 3d human pose
  estimation in video.
\newblock In {\em Proceedings of the IEEE/CVF Conference on Computer Vision and
  Pattern Recognition}, pages 13232--13242, 2022.

\bibitem{zhao2022graformer2}
Weixi Zhao, Weiqiang Wang, and Yunjie Tian.
\newblock Graformer: Graph-oriented transformer for 3d pose estimation.
\newblock In {\em Proceedings of the IEEE/CVF Conference on Computer Vision and
  Pattern Recognition}, pages 20438--20447, 2022.

\bibitem{Zhao_2023_tip_fsc}
Yunqing Zhao and Ngai-Man Cheung.
\newblock Fs-ban: Born-again networks for domain generalization few-shot
  classification.
\newblock {\em IEEE Transactions on Image Processing}, 2023.

\bibitem{zheng20213d2}
Ce Zheng, Sijie Zhu, Matias Mendieta, Taojiannan Yang, Chen Chen, and Zhengming
  Ding.
\newblock 3d human pose estimation with spatial and temporal transformers.
\newblock In {\em Proceedings of the IEEE/CVF International Conference on
  Computer Vision}, pages 11656--11665, 2021.

\end{thebibliography}
}

\end{document}